\PassOptionsToPackage{table}{xcolor}
\documentclass[]{bytedance_seed}

\usepackage[toc,page,header]{appendix}
\usepackage{booktabs}
\usepackage{amsmath}
\usepackage{amssymb}
\usepackage{amsfonts}
\usepackage{nicefrac}
\usepackage{enumitem}
\usepackage{graphicx}
\usepackage{multirow}
\usepackage{makecell}
\usepackage{adjustbox}
\usepackage{xspace}
\usepackage{pifont}
\usepackage{algorithm}
\usepackage{algpseudocode}
\usepackage{amsthm}
\usepackage{float}

\usepackage{tikz}
\usetikzlibrary{arrows.meta, positioning, calc, shapes.geometric, shapes.misc, decorations.pathmorphing, decorations.markings, fit, backgrounds, patterns, shadings, matrix}
\usepackage{pgfplots}
\pgfplotsset{compat=1.18, /pgf/number format/use comma=false}

\definecolor{vefblue}{RGB}{30,90,170}
\definecolor{vefred}{RGB}{200,60,60}
\definecolor{vefgreen}{RGB}{40,130,90}
\definecolor{vefgrey}{RGB}{120,120,120}
\definecolor{vefyellow}{RGB}{220,180,40}
\definecolor{vefpurple}{RGB}{120,80,170}

\definecolor{vefblue1}{RGB}{49,89,180}
\definecolor{vefblue2}{RGB}{80,135,242}
\definecolor{vefblue3}{RGB}{86,188,199}
\definecolor{vefblue4}{RGB}{148,228,221}

\hypersetup{
  colorlinks=true,
  linkcolor=vefblue3,
  citecolor=vefblue3,
  urlcolor=vefblue3,
}

\colorlet{vefoursrow}{vefpurple!8}
\colorlet{vefbaserow}{vefyellow!12}

\definecolor{vefaccent}{RGB}{75,85,170}
\newcommand{\name}{\textsl{\textbf{\textcolor{vefblue2}{D}\textcolor{vefblue3}{ance}\textcolor{vefblue4}{OPD}}}\xspace}

\renewcommand{\checkmark}{\textcolor{vefblue3}{\ding{51}}}

\title{DanceOPD: On-Policy Generative Field Distillation}

\author[1,2,\ddagger]{Wei Zhou}
\author[1]{Xiongwei Zhu}
\author[1]{Zelin Xu}
\author[1]{Bo Dong}
\author[1]{Lixue Gong}
\author[3]{Yongyuan Liang}
\author[4]{Meng Chu}
\author[2]{Leigang Qu}
\author[2]{Lingdong Kong}
\author[1,\dagger]{Wei Liu}
\author[2]{Tat-Seng Chua}

\affiliation[1]{ByteDance Seed}
\affiliation[2]{NUS}
\affiliation[3]{UMD}
\affiliation[4]{HKUST}

\contribution[\ddagger]{Work done at ByteDance Seed}
\contribution[\dagger]{Corresponding author}

\abstract{Modern image generation demands a single model that unifies diverse capabilities, including text-to-image (T2I), local editing, and global editing. However, these capabilities are rarely naturally aligned and often conflict. 
For instance, editing tends to degrade T2I performance, while global and local editing interfere with each other. Consequently, effectively composing these capabilities has become a central challenge for image generation model training. To tackle this, we introduce \name{}, an on-policy generative field distillation framework for flow-matching models that routes each sample to one capability field, queries one low-noise student-induced state, and trains with a simple velocity MSE objective. With each capability source defined as a velocity field over the shared flow state space, the student learns from fields queried on its own rollout states to compose expert capabilities. This formulation also absorbs operator-defined fields such as classifier-free guidance. Comprehensive experiments on T2I, editing, realism-field absorption, and CFG absorption show that our approach improves multi-capability composition, strengthening target capabilities while preserving anchor generation quality. We believe this work establishes a practical route for generative field distillation in flow-matching models.
}

\checkdata[Project Page]{\href{https://DanceOPD.github.io}{\textbf{DanceOPD.github.io}}~~~~{{\fontsize{7.92}{10}\textbf{GitHub Repo:}}} \href{https://github.com/worldbench/DanceOPD}{\textbf{github.com/worldbench/DanceOPD}}}
\correspondence{Wei Liu (E-mail: \url{liuwei.jikun@bytedance.com})}
\date{June 26, 2026}

\begin{document}

\maketitle
\vspace{-0.6cm}

\begin{figure}[H]
\centering
\vspace{-0.4cm}
\resizebox{\linewidth}{0.35\linewidth}{\includegraphics{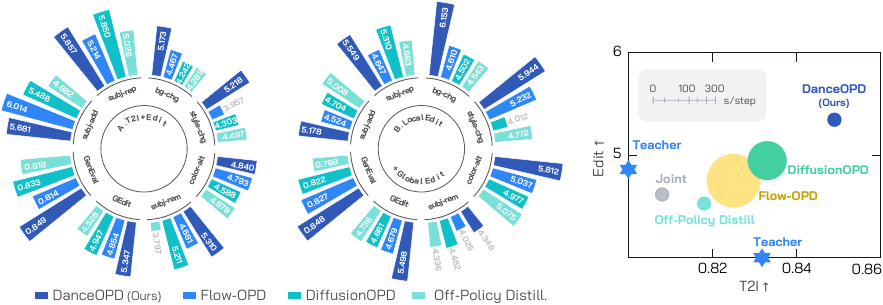}}
\vspace{-0.55cm}
\caption{\textbf{\name{} Improves Multi-Capability Composition.}
\textbf{Left:} Per-metric performance for the two composition settings, compared with representative baselines.
\textbf{Right:} Editing$\times$T2I capability space for their composition, with marker size denoting per-step training cost. DanceOPD achieves a better composition while incurring a lower cost.
}
\label{fig:overview}
\vspace{-0.35cm}
\end{figure}

\clearpage\clearpage
\begin{figure}[H]
\centering
\vspace{-0.5cm}
\resizebox{0.99\linewidth}{!}{\includegraphics{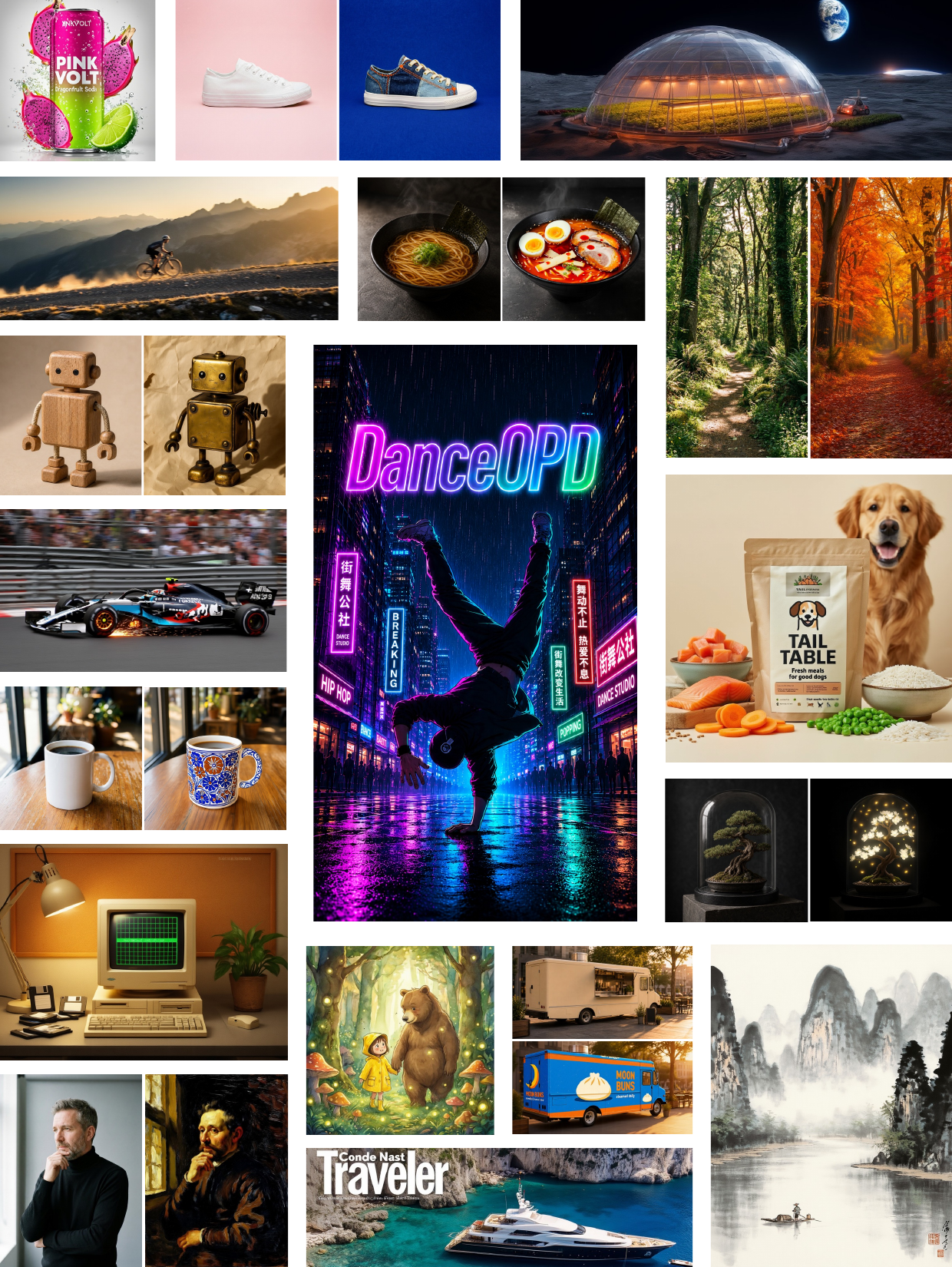}}
\vspace{-0.15cm}
\caption{\textbf{Qualitative Examples from \name{}.} After training with DanceOPD, the resulting student supports diverse text-to-image and editing behaviors while retaining strong original generative capabilities.}
\label{fig:viz}
\end{figure}

\clearpage\clearpage
\setcounter{tocdepth}{3}
\tableofcontents

\clearpage\clearpage
\section{Introduction}
\label{sec:introduction}

Modern image generation has rapidly advanced through flow matching models~\cite{ho2020denoising,rombach2022high,peebles2023scalable,esser2024scaling,cai2025z,podell2024sdxl}. A single deployed model is increasingly expected to jointly support diverse capabilities, including text-to-image generation (T2I), local editing, and global editing~\cite{brooks2023instructpix2pix,yu2025anyedit,chow2025editmgt}. These capabilities, however, are not naturally compatible. T2I rewards open-ended visual quality and prompt following; local editing requires preserving the input while applying precise changes; and global editing intentionally changes broad appearance statistics such as style, color, or layout. Naively optimizing them together often leads to capability interference: editing can erode T2I performance, while local and global editing can pull the model toward conflicting preservation and transformation behaviors. Consequently, how to effectively compose these capabilities has become a central challenge for the image generation model community.

Existing capability-combination paradigms only partially address this goal. Data mixing or joint training tends to dilute capability-specific supervision and can suffer from multi-task gradient conflict~\cite{ilharco2022editing,wortsman2022model,matena2022merging,chen2018gradnorm,sener2018multi,javaloy2021rotograd}; parameter-space merging and adapter composition typically yield compromise solutions~\cite{ainsworth2022git,yadav2023ties,gal2022image,ruiz2023dreambooth,pfeiffer2021adapterfusion}; and inference-time score composition leaves the composition external to the deployed student~\cite{liu2022compositional,du2023reduce,bar2023multidiffusion}. While useful, these approaches do not directly resolve the state-dependent question: 

\textit{How should a model effectively compose multiple generative capabilities?}

Building on recent formulations that interpret task abilities as velocity fields in flow-matching image generators~\cite{lipman2022flow,liu2022flow,albergo2025stochastic,tong2023improving,dao2023flow}, we adopt a field-based perspective in which each frozen capability source defines a velocity field over a shared generative state space. Under this view, capability composition reduces to a field-query problem with three coupled choices: which capability field should supervise a given sample, where in the state space the field should be queried, and how many states from the student rollout should be used for supervision. This formulation aligns naturally with on-policy distillation (OPD)~\cite{ross2011reduction,gu2024minillm,agarwal2024policy,jia2026asymmetric}, in which the teacher supervises states produced by the current student rather than fixed offline trajectories.

These three choices directly determine the main alignment challenges in multi-capability distillation. For the choice of target field, linearly combining several frozen fields within a single sample target can produce a supervision direction that does not correspond to any well-defined capability query, causing target-field ambiguity and losing the semantic identity of each capability. For the choice of query state, evaluating the field on data states, teacher trajectories, or other off-policy states leaves the student under-supervised on the states it actually visits during generation, causing state-distribution mismatch and a gap between training and inference. For the choice of trajectory supervision, drawing dense targets from multiple states along the same rollout overcounts highly correlated supervision signals, since these states share the same prompt, noise seed, student dynamics, and path history. This causes trajectory-query correlation and can bias the resulting gradient.

Guided by these diagnoses, we introduce \name{}, an on-policy generative field distillation framework whose three design choices correspond to the three challenges above.
\begin{itemize}
    \item To resolve target-field ambiguity, our method performs hard-routed sample-wise field matching: each sample is dispatched to exactly one frozen capability field, preserving its semantic identity.

    \item To resolve state-distribution mismatch, the routed field is queried on a stop-gradient state drawn from the current student rollout, aligning supervision with the student's own visitation distribution.

    \item To resolve trajectory-query correlation, we use a single semantic-side low-noise query per sample, located in the regime where capability-specific information is most concentrated, thereby avoiding correlated within-trajectory samples.
\end{itemize}

The student is updated by a plain velocity MSE loss, which is the natural local regression objective for deterministic velocity fields: under a local Gaussian transition view, KL-style field matching reduces to a weighted MSE form, as derived in Appendix~Sec.~\ref{app:theory}. The same formulation subsumes operator-defined fields; classifier-free guidance, in particular, can be cast as a guided velocity field and absorbed into the student under the identical objective~\cite{ho2022classifier}.

We evaluate our approach across two capability-composition settings and two field-absorption cases: T2I and editing composition, local and global editing composition, realism-field absorption with base T2I preservation, and CFG absorption. The main results show that hard-routed on-policy field matching improves multi-capability composition, strengthening target edit or realism-oriented fields while preserving base generation quality. 
In T2I and editing composition, \name{} improves GEditBench over the best reproduced OPD baseline by $8.1\%$ and over the edit source by $8.5\%$, while slightly exceeding the T2I source on GenEval. 
In local and global edit composition, our model improves over the best competing composition baseline by $16.1\%$ and over the local edit source by $7.9\%$, with GenEval above all compared composition baselines. 

For realism-field absorption, our model improves the realism reward over off-policy distillation by $9.9\%$ and closes $85.3\%$ of the student-to-teacher reward gap while maintaining the T2I score within $0.1\%$ of off-policy distillation and above the student anchor by $7.6\%$. For CFG absorption, the best measured composition improves over train-only absorption by $7.6\%$ and over eval-only CFG by $1.4\%$, while over-guided composition drops substantially. The diagnostic studies further support the method design: hard routing improves over soft all-teacher mixing by $15.2\%$ under MSE and $10.6\%$ under KL; semantic-side low-noise queries improve over median- and high-noise queries by $23.7\%$ and $19.5\%$; dense same-step accumulation drops by $22.8\%$, and SDE-style decorrelation partially rescues the stress case with an $18.4\%$ gain but remains $8.6\%$ below the single-query default; and plain velocity MSE improves over the main weighted alternatives by $2.8\%$ to $4.5\%$.

Our contributions can be summarized as follows:
\begin{itemize}[left=1pt]
    \item We formulate multi-capability image generation as on-policy generative field distillation and introduce \name{}, a hard-routed, semantic-side velocity-matching method on student-visited states.
    \item We identify three query-induced alignment challenges, including target-field ambiguity, state-distribution mismatch, and trajectory-query correlation, and show how they motivate our design.
    \item Experiments across T2I and editing composition, realism-field absorption, and CFG absorption, together with comprehensive ablations, confirm the effectiveness of DanceOPD.
\end{itemize}

\section{Related Work}
\label{sec:related}

\subsection{Multi-Capability Composition}
Modern image generation increasingly requires a single deployed model to support multiple capabilities, including open-ended text-to-image generation~\cite{rombach2022high,peebles2023scalable,cai2025z}, local editing that preserves the source image while applying targeted changes~\cite{brooks2023instructpix2pix,yu2025anyedit,bai2024humanedit}, global editing that changes broad appearance, layout, or style statistics~\cite{chow2025editmgt,chow2026editmgt_workshop}, and style-specialized or personalized generation~\cite{gal2022image,ruiz2023dreambooth,shah2024ziplora}. These capabilities are not naturally compatible: text-to-image generation emphasizes open-ended visual quality and prompt following, local editing emphasizes source preservation under precise changes, and global or style-oriented editing emphasizes broader transformation. Therefore, multi-capability composition is not merely a matter of covering more generation and editing methods; it asks how one deployed student can strengthen a target capability while preserving the anchor capability.
Existing multi-capability composition methods often incur degradation in individual capabilities when combining these abilities. Data mixing or joint training can dilute capability-specific supervision and expose gradient conflict~\cite{chen2018gradnorm,sener2018multi}; parameter-space merging often yields compromise solutions~\cite{yadav2023ties,yang2024adamerging}, as do personalization and adapter composition~\cite{kumari2023multi,shah2024ziplora} and adapter-fusion analyses~\cite{mahabadi2021parameter,chen2025empirical}; and inference-time score composition leaves the composition outside the deployed student~\cite{liu2022compositional,bar2023multidiffusion}. In contrast, \name{} treats each frozen capability source as a velocity field~\cite{lipman2022flow,liu2022flow} and learns a single student through hard-routed on-policy field matching, thereby combining the strengths of different teacher models and even achieving performance beyond individual teachers, as shown in Fig.~\ref{fig:overview}. Table~\ref{tab:positioning} summarizes how this positioning differs from prior on-policy distillation methods.

\begin{table*}[t]
  \centering
  \setlength{\tabcolsep}{3pt}
  \caption{\textbf{Comparison with OPD Methods.} To our knowledge, \name{} is the only method that combines flow-matching OPD, multi-capability composition, design-space analysis, and functional field absorption.}
  \vspace{-0.2cm}
  \label{tab:positioning}
  \adjustbox{max width=\textwidth}{%
  \begin{tabular}{llcc|cccc}
    \toprule
    \textbf{Method} & \textbf{Domain} & \textbf{Teacher Signal} &
    \textbf{Objective} & \textbf{FM-OPD} & \textbf{Multi-Cap.} &
    \textbf{Design Study} & \textbf{Func. Absorp.}\\
    \midrule\midrule
    MiniLLM~\cite{gu2024minillm} & LLM & logits & reverse KL
      & -- & -- & -- & --\\
    GKD~\cite{agarwal2024policy} & LLM & logits & forward KL
      & -- & -- & -- & --\\
    AOPD~\cite{jia2026asymmetric} & LLM & logits and top-$K$ & asymmetric KL
      & -- & -- & -- & --\\
    \midrule
    G-OPD~\cite{yang2026learning} & LLM & scalar feedback & policy optimization
      & -- & $\circ$ & -- & --\\
    StableOPD~\cite{luo2026demystifying} & LLM & scalar reward & PPO with KL anchor
      & -- & $\circ$ & -- & --\\
    ROPD~\cite{fang2026rubric} & LLM & rubric reward & reward optimization
      & -- & $\circ$ & -- & --\\
    \midrule
    DiffusionOPD~\cite{li2026diffusionopd} & Flow & velocity field & KL or MSE-style
      & \checkmark & $\circ$ & $\circ$ & --\\
    D-OPSD~\cite{jiang2026d} & Diffusion & predicted distribution & self-distillation
      & $\circ$ & -- & -- & --\\
    Flow-OPD~\cite{fang2026flow} & Flow & dense scalar reward & PPO clip-min
      & \checkmark & task-routed & -- & --
      \\
      \midrule
    \rowcolor{vefblue3!9}
    \textbf{\ding{72}~\name{}} & Flow & routed velocity field & MSE
      & \checkmark & \checkmark & \checkmark & \checkmark\\
    \bottomrule
  \end{tabular}%
  }
\end{table*}

\subsection{On-Policy Distillation}
On-policy distillation addresses the mismatch between fixed training states and states induced by the current student. It differs from standard knowledge distillation and diffusion distillation, which mainly target model compression, sampler compression, trajectory training, consistency training, or distribution matching~\cite{hinton2015distilling,salimans2022progressive,song2023consistency,luo2023latent,yin2024one,sauer2024fast,kim2024consistency}. In language models, KL-based OPD methods match teacher token distributions on student-generated sequences~\cite{gu2024minillm,agarwal2024policy,jia2026asymmetric}, while reward-based variants optimize scalar feedback with policy-gradient or PPO-style objectives~\cite{yang2026learning,luo2026demystifying,fang2026rubric}. Recent work further shows that rollout position affects OPD stability and efficiency~\cite{ziheng2026moreearlystoppingrollout,kong2026ai}, suggesting that the location of teacher supervision along a student rollout is itself an important design choice.

Recent flow-model OPD methods instantiate related ideas for generative post-training. DiffusionOPD performs on-policy velocity matching, D-OPSD studies on-policy self-distillation, and Flow-OPD uses dense reward optimization with PPO-style clipping~\cite{li2026diffusionopd,jiang2026d,fang2026flow}. Our focus is complementary: we study how T2I, editing, realism, and guidance-related generative capabilities can be composed through on-policy field distillation in flow-matching models. This requires deciding which teacher field supervises each sample, where the field should be queried, and how many rollout states should contribute supervision. We therefore compare KL-style objectives, teacher routing, query states, and dense versus single-query supervision, and find that one low-noise semantic query on the student's rollout is an effective and substantially cheaper default. We also evaluate realism-field absorption, CFG absorption, and mismatched training rollout and evaluation sampler step counts, which reflect practical post-training pipelines.

\subsection{Generative Field Distillation}
Flow-matching formulations view generation as learning a velocity field over a continuous state space, building on diffusion and score-based generative modeling foundations~\cite{ho2020denoising,lipman2022flow,liu2022flow,albergo2025stochastic,tong2023improving,dao2023flow}. This field view is useful for capability composition because a frozen T2I model, edit model, realism-oriented model, or guidance operator can all be treated as sources of local velocity supervision on a shared state space. Prior work has explored related ingredients: score or field composition combines multiple generative signals at inference time~\cite{du2023reduce,bar2023multidiffusion}, multi-task optimization studies how to handle conflicting gradients during joint training~\cite{chen2018gradnorm,sener2018multi,yu2020gradient,liu2021conflict,javaloy2021rotograd}, and on-policy imitation or distillation emphasizes supervision on states visited by the current student~\cite{ross2011reduction,li2026diffusionopd,fang2026flow}. These directions motivate our formulation but do not specify how multiple capability fields should be queried during distillation. \name{} addresses this missing design problem with hard field routing, student-induced query states, and one semantic-side low-noise query.

\section{Approach}
\label{sec:approach}

\begin{figure}[t]
  \centering
  \includegraphics[width=\linewidth]{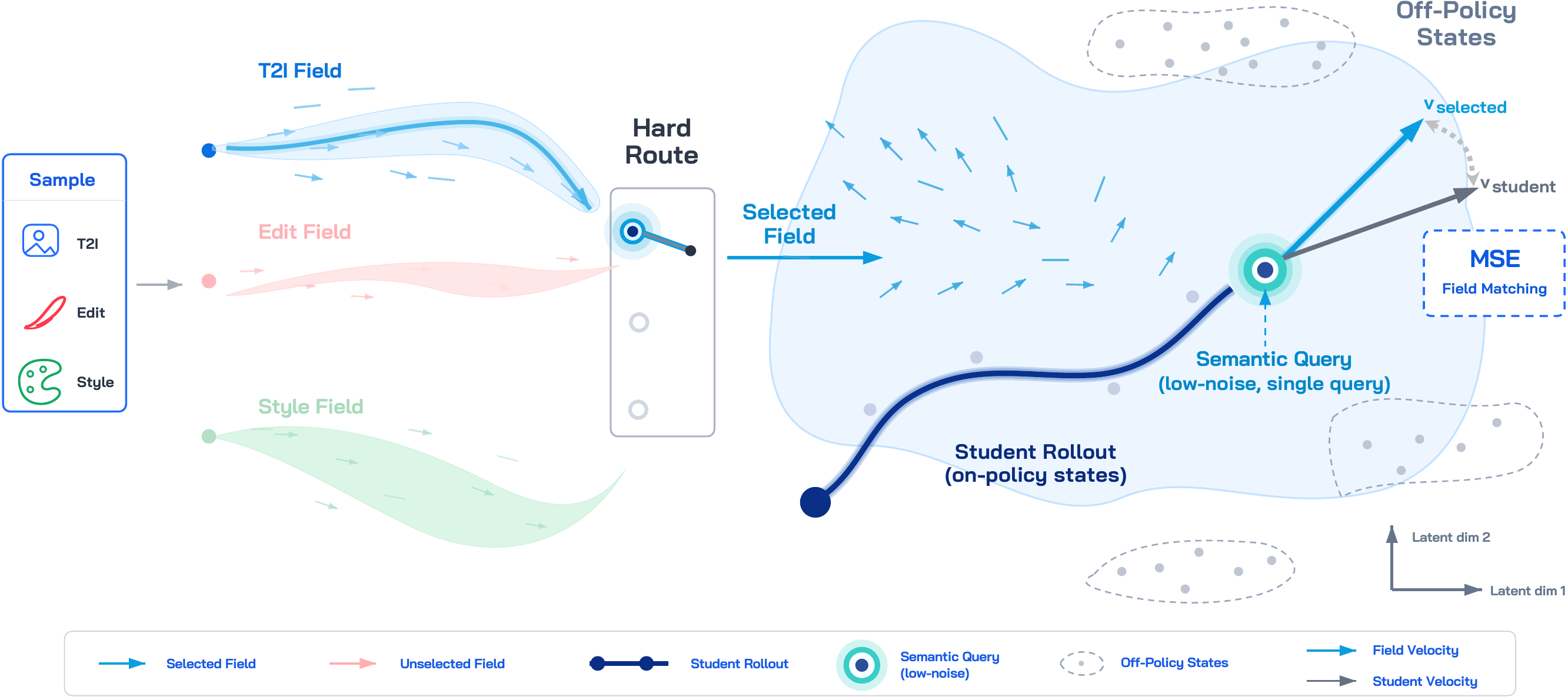}
  \vspace{-0.5cm}
  \caption{\textbf{Conceptual Illustration of \name{}.} For each sample, DanceOPD hard-routes supervision to one capability field, queries that field at a single low-noise state on the current student rollout rather than off-policy states, and aligns the student velocity to the selected field velocity with MSE.}
  \label{fig:method-query}
\end{figure}

We formulate multi-capability image generation as \emph{on-policy generative field distillation}. Given several frozen capability sources, our goal is to train one student to query and imitate these sources as velocity fields on a shared generative state space~\cite{lipman2022flow,liu2022flow,albergo2025stochastic}, rather than combining them through static parameter interpolation or data-ratio tuning~\cite{ilharco2022editing,wortsman2022model,yadav2023ties}. This formulation turns capability composition into a field-query problem with three coupled choices: which capability field should supervise a sample, where in the state space the field should be queried, and how many states from the student rollout should be used for supervision. Fig.~\ref{fig:method-query} summarizes the resulting training query. Sec.~\ref{sec:hard-routing} to Sec.~\ref{sec:query-design} instantiate these choices through hard-routed field selection, on-policy student-state querying, and semantic-side single-query supervision, addressing target-field ambiguity, state-distribution mismatch, and trajectory-query correlation, respectively; Sec.~\ref{sec:objective-design} then gives the local matching objective.

\subsection{Preliminary}
\label{sec:composition-target}

Assume we are given $M\ge2$ frozen capability sources that are defined on the same generative state space and expose compatible velocity predictions. The sources may correspond to different tasks or behaviors, such as text-to-image, image editing, local attribute editing, global editing, or style-specialized generation. We denote the trainable student by $v_\theta$, whose parameters are updated during distillation, and the frozen capability sources by $\{v_m\}_{m=1}^{M}$. For each source, $\mathcal{D}_m$ denotes the corresponding route-specific training distribution, $x$ denotes the route-specific data payload, and $p_T$ denotes the initial-noise distribution. This source-selection view is related to conditional expert routing and multi-task gating~\cite{jacobs1991adaptive,shazeer2017outrageously,ma2018modeling}.

A naive unified model often suffers from \emph{capability dilution}: joint training averages supervision signals, weight merging assumes approximate parameter-space linearity, and data-ratio tuning merely moves the model along a trade-off curve~\cite{chen2018gradnorm,sener2018multi,yu2020gradient,liu2021conflict}. Our goal is different. We define the desired composition outcome operationally: under the reported metrics, a single model should improve the target capability while preserving the anchor capability. This does not claim global Pareto optimality over all possible training mixtures, but specifies the target region for capability composition.

We view each frozen source as defining a capability-specific velocity field:
\begin{equation}
  v_m(z_t,t,c), \qquad m\in\{1,\ldots,M\},
  \label{eq:cap-field}
\end{equation}
where $z_t$ is a flow state at time $t$ and $c$ denotes the conditioning information, such as a text prompt, source image, edit instruction, or style condition. This abstraction is independent of how the source is obtained: a source can be a pretrained model, a fine-tuned model, or a task-specialized branch.

Under this view, capability composition becomes a field-alignment problem: which capability field should the student imitate, and at which states should the field be queried? This yields three design questions that structure the rest of the method: which field supervises each sample, where the field is queried, and how many trajectory states are used for supervision.

\subsection{Hard-Routed Sample-Wise Field Matching}
\label{sec:hard-routing}

The first design question is which field should supervise each sample. A naive alternative is to average several fields inside the same sample target, as in soft multi-teacher or knowledge-amalgamation variants~\cite{hinton2015distilling,shen2019customizing}. However, such an averaged target may no longer correspond to any well-defined capability query, especially when the fields encode different tasks or visual behaviors. To keep the target semantically meaningful, we route each training sample to exactly one capability field. A sample drawn from the text-to-image partition queries the T2I field; an edit sample queries the edit field; a local-edit or style-edit sample queries the corresponding field. Let $\pi(m)$ denote the route probability over active capability buckets. Formally, this is:
\begin{equation}
  m\sim \pi(m), \qquad (x,c)\sim \mathcal{D}_m,
  \label{eq:sample-route}
\end{equation}
We do not tune $\pi$; unless otherwise stated, it is uniform over the active capability buckets, so two-bucket composition uses a $1{:}1$ route ratio and the three-bucket diagnostics use a $1{:}1{:}1$ route ratio.
We then define the routed target field as:
\begin{equation}
  u_m(z,t,c)=v_m(z,t,c).
  \label{eq:routed-field}
\end{equation}

This routing preserves the semantic identity of each capability query, in contrast to multi-teacher distillation schemes that combine teacher signals before or within a student update~\cite{you2017learning,fukuda2017efficient}. It avoids injecting unrelated capability directions into a single sample target, and it avoids forcing several potentially conflicting fields to be averaged inside one sample-level supervision signal. Capability composition is therefore achieved statistically across routed updates, while each individual query remains semantically well-defined.

This design is also consistent with observations in multi-task optimization: when task gradients are poorly aligned, averaging them in the same optimizer step can produce a compromise direction rather than progress for every task~\cite{chen2018gradnorm,sener2018multi,yu2020gradient,liu2021conflict,javaloy2021rotograd}. Our method, therefore, uses hard-routed sample-wise field matching as the default capability-composition estimator and studies same-step multi-field accumulation only as an ablation.

\subsection{On-Policy Field Querying}
\label{sec:onpolicy-query}

The second design question is where the routed field should be queried. A student that composes several capabilities generally follows a trajectory different from any individual frozen source. Therefore, aligning fields only on data states or on teacher-induced trajectories can leave a mismatch between training and testing: at inference, the student visits states $z_t^\theta$ that are not necessarily covered by the off-policy query distribution.

\name{} instead queries the capability fields on the student's own rollout. Let:
\begin{equation}
  z_{0:T}^{\theta}=\mathrm{Rollout}(v_\theta; z_T,c), \qquad z_T\sim p_T,
  \label{eq:student-rollout}
\end{equation}
be the trajectory induced by the current student from initial noise $z_T$. We sample a semantic coordinate $s$ and its physical timestep $t=t(s)$, set the query state to $\bar z_t=\mathrm{sg}(z_t^\theta)$, query the frozen capability field as $v_m(\bar z_t,t,c)$, and train the student to match that field locally. This is the flow-model counterpart of on-policy distillation: the frozen source supervises the student on the distribution induced by the student itself, addressing the same covariate-shift issue that motivates on-policy imitation learning and recent OPD methods~\cite{ross2011reduction,agarwal2024policy,li2026diffusionopd,fang2026flow}. The stop-gradient query means that the update differentiates the local velocity prediction, not the entire rollout solver. Appendix~Sec.~\ref{app:onpolicy-sg} gives the corresponding estimator and mismatch bound.

\subsection{Semantic-Side Single Query}
\label{sec:query-design}

The third design question is how many states should be queried from the student trajectory. A natural extension is to query many states from the same rollout. However, dense trajectory queries can fail because multiple states from one rollout are correlated: they share the same initial noise, prompt, conditioning, student dynamics, and path history. Thus, their gradients are not independent, and adding more states does not necessarily provide more independent supervision, echoing multi-task optimization settings where gradient aggregation itself can become the bottleneck~\cite{chen2020just,liu2021towards,navon2022multi}.

These considerations motivate a semantic-side single query. Let $s\in[0,1]$ be a normalized semantic-side rollout coordinate, and let $q_{\mathrm{sem}}$ be a query distribution biased toward the low-noise side of the trajectory. Rather than supervising the whole trajectory, \name{} samples one high-information state:
\begin{equation}
  K=1, \qquad s\sim q_{\mathrm{sem}}(s),\quad t=t(s),
  \label{eq:semantic-query}
\end{equation}
High-noise states can contain coarse structure, but they are often dominated by generic denoising and have a lower density of capability-specific signal. Low-noise states are closer to the final image and concentrate style, aesthetics, local attributes, and task-specific edit information. This view is consistent with prior diffusion editing and distillation studies showing that timestep and noise-schedule choices strongly affect the fidelity and editability trade-off and that semantic-related timesteps are especially important for image manipulation~\cite{meng2021sdedit,hertz2022prompt,wang2023not,chen2024tino,lin2024schedule}. Thus, a single low-$t$ query provides a high signal-to-noise capability supervision while avoiding correlated trajectory samples.

Dense-query variants are useful as diagnostics. If dense-query degradation is partly caused by trajectory correlation, then decorrelating the rollout should partially mitigate this failure in dense-query stress cases. We test this prediction with stochastic rollout noise, which reduces query correlation. The detailed SDE formulation and the correlation-decay argument are deferred to the appendix.

\begin{algorithm}[t]
\caption{\textbf{One \name{} Training Step.} A sample is routed to one capability field, queried on a semantic-side student state, and updated by velocity matching.}
\label{alg:danceopd}
\small
\begin{algorithmic}[1]
\Require Frozen capability fields $\{v_m\}_{m=1}^M$, student $v_\theta$, route probabilities $\pi$, query distribution $q_{\mathrm{sem}}$ over semantic coordinate $s$.
\Ensure Updated student parameters $\theta$.
\Statex \textbf{\textcolor{vefblue2}{A.} Route One Capability Query}
\State $m \sim \pi(m),\quad (x,c)\sim\mathcal{D}_m$ \Comment{preserve sample identity}
\Statex \textbf{\textcolor{vefblue3}{B.} Query On The Student Trajectory}
\State $z_T\sim p_T,\quad z_{0:T}^{\theta}\gets \mathrm{Rollout}(v_\theta;z_T,c)$
\State $s\sim q_{\mathrm{sem}}(s),\quad t\gets t(s),\quad \bar z_t\gets \mathrm{sg}(z_t^\theta)$ \Comment{one semantic-side state}
\State $u \gets v_m(\bar z_t,t,c)$ \Comment{frozen routed field}
\Statex \textbf{\textcolor{vefblue4}{C.} Match The Local Velocity Field}
\State $\mathcal{L}\gets \bigl\|v_\theta(\bar z_t,t,c)-u\bigr\|_2^2$
\State $\theta\gets \mathrm{OptStep}\!\left(\theta,\nabla_\theta\mathcal{L}\right)$
\end{algorithmic}
\end{algorithm}

\subsection{Objective Design}
\label{sec:objective-design}

Once the routed field, student-induced query state, and semantic-side query time are fixed, the remaining choice is the local matching objective. Our default objective is plain velocity MSE on the routed, on-policy query. With the stop-gradient query state $\bar z_t=\mathrm{sg}(z_t^\theta)$, we optimize:
\begin{equation}
  \mathcal{L}_{\mathrm{DanceOPD}}
  =
  \mathbb{E}_{m\sim\pi,\,(x,c)\sim\mathcal{D}_m,\,z_T\sim p_T,\,s\sim q_{\mathrm{sem}}}
  \left[
    \left\|v_\theta(\bar z_t,t,c)-v_m(\bar z_t,t,c)\right\|_2^2
  \right], \qquad t=t(s).
  \label{eq:danceopd-mse}
\end{equation}
The target is a deterministic velocity field, making MSE the natural regression objective under an isotropic Gaussian velocity-likelihood view~\cite{lipman2022flow,ho2020denoising}. KL-style velocity matching reduces to a weighted MSE form under the same view; we provide the derivation in Appendix~Sec.~\ref{app:kl-mse}. This statement is limited to the local velocity-matching subproblem and should not be interpreted as an information-theoretic ceiling on downstream task metrics. Empirically, Sec.~\ref{sec:exp-ablation} shows that plain MSE is the most stable and best-performing objective among the tested alternatives, including KL weighting, timestep-weighted objectives, and score-style and consistency-style surrogates.

The same formulation also applies to operator-defined fields. For example, let $v_{\emptyset}$ and $v_{\mathrm{cond}}$ denote the unconditional and conditional velocity predictions. Classifier-free guidance with guidance scale $\alpha$ defines a guided velocity field, that is:
\begin{equation}
  v_{\alpha}(z_t,t,c)
  = v_{\emptyset}(z_t,t)
  + \alpha\bigl(v_{\mathrm{cond}}(z_t,t,c)-v_{\emptyset}(z_t,t)\bigr),
  \label{eq:cfg-field}
\end{equation}
which can be treated as an additional capability field and absorbed into the student by the same MSE objective~\cite{ho2022classifier}.

Fig.~\ref{fig:editing_3} further illustrates how the same field-matching objective handles material, environment, and lighting edits. In the apple example, \name{} transforms the red apple into a clear crystal-like object while keeping the leaf, tabletop, and studio lighting layout. In the piano example, DanceOPD introduces an underwater environment with caustic light and bubbles while retaining the grand-piano structure. In the leather-armchair example, the chair geometry and leather appearance remain visible under a stronger cinematic spotlight, whereas several baselines either over-expose the object or make the scene too dark to preserve the chair details. These cases support the role of routed, semantic-side supervision: the edit field can change the requested visual factor without forcing unrelated changes to the source object or scene layout.
\section{Experiments}
\label{sec:experiments}

We evaluate \name{} primarily on Z-Image~\cite{cai2025z} for composing frozen capability fields into one student, and use SD3.5-M for the realism-field absorption setting, with implementation details provided in the appendix. The experiments are organized around three questions: whether a single student can compose heterogeneous capability fields without collapsing into a compromise between them; which field-routing and query choices are needed when multiple capability fields supervise the same student; and how the main training choices, including student-induced query states, semantic-side single querying, rollout discretization, initialization, and plain velocity MSE, affect performance. Sec.~\ref{sec:main-results} reports the main results for capability composition and field absorption under a unified protocol, Sec.~\ref{sec:multi-teacher-ext} studies the multi-teacher extension and routing diagnostics, and Sec.~\ref{sec:exp-ablation} ablates query and objective design. Across all settings, the main finding is that DanceOPD improves the target capability while preserving the anchor capability, whereas joint training, weight merging, soft teacher mixing, and dense same-step supervision tend to reintroduce capability interference.

\subsection{Main Results}
\label{sec:main-results}

To evaluate the effectiveness of \name{}, we consider four settings that cover both capability composition and field absorption: T2I and editing composition, local and global editing composition, realism-field absorption with base T2I preservation, and CFG absorption. For the two editing-related composition settings, we use GenEval to measure general text-to-image generation ability and GEditBench to measure general editing ability~\cite{ghosh2023geneval,liu2025step1x}. For realism-field and CFG absorption, we use diagnostic metrics matched to the absorbed target fields while also monitoring preservation of the anchor generation capability. Across these settings, DanceOPD consistently strengthens the target capability while preserving the anchor capability, showing that routed on-policy field matching can compose and absorb generative fields without collapsing into a simple compromise between sources.

\begin{figure}[H]
    \centering
    \includegraphics[width=\textwidth]{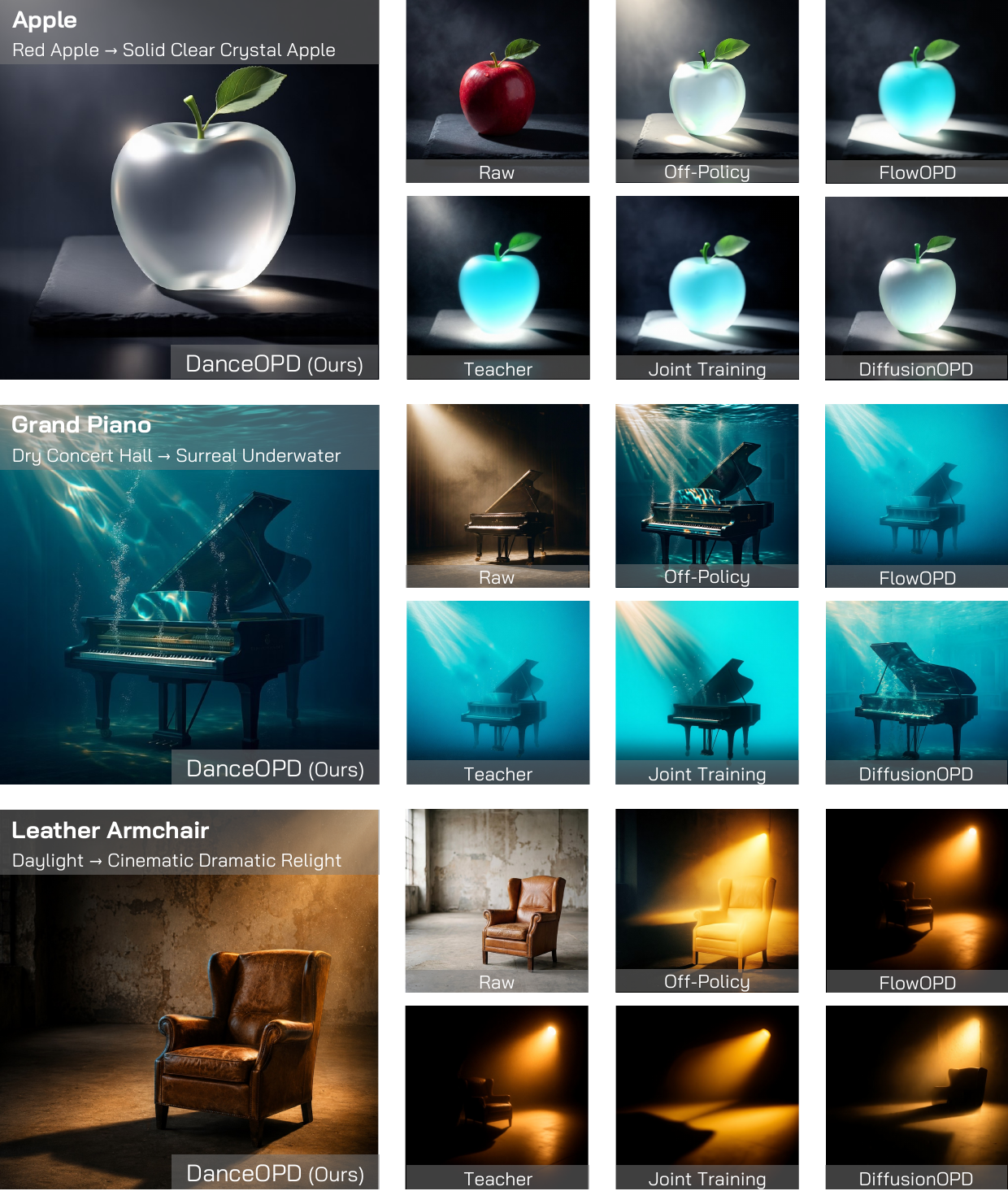}
    \vspace{-0.5cm}
    \caption{\textbf{T2I and Edit Composition: Additional Edit Cases.} DanceOPD better balances target attribute changes with content preservation across material, lighting, and style edits.}
    \label{fig:editing_3}
    \vspace{-1.2cm}
\end{figure}

\begin{figure}[H]
    \centering
    \vspace{-0.9cm}
    \includegraphics[width=\textwidth]{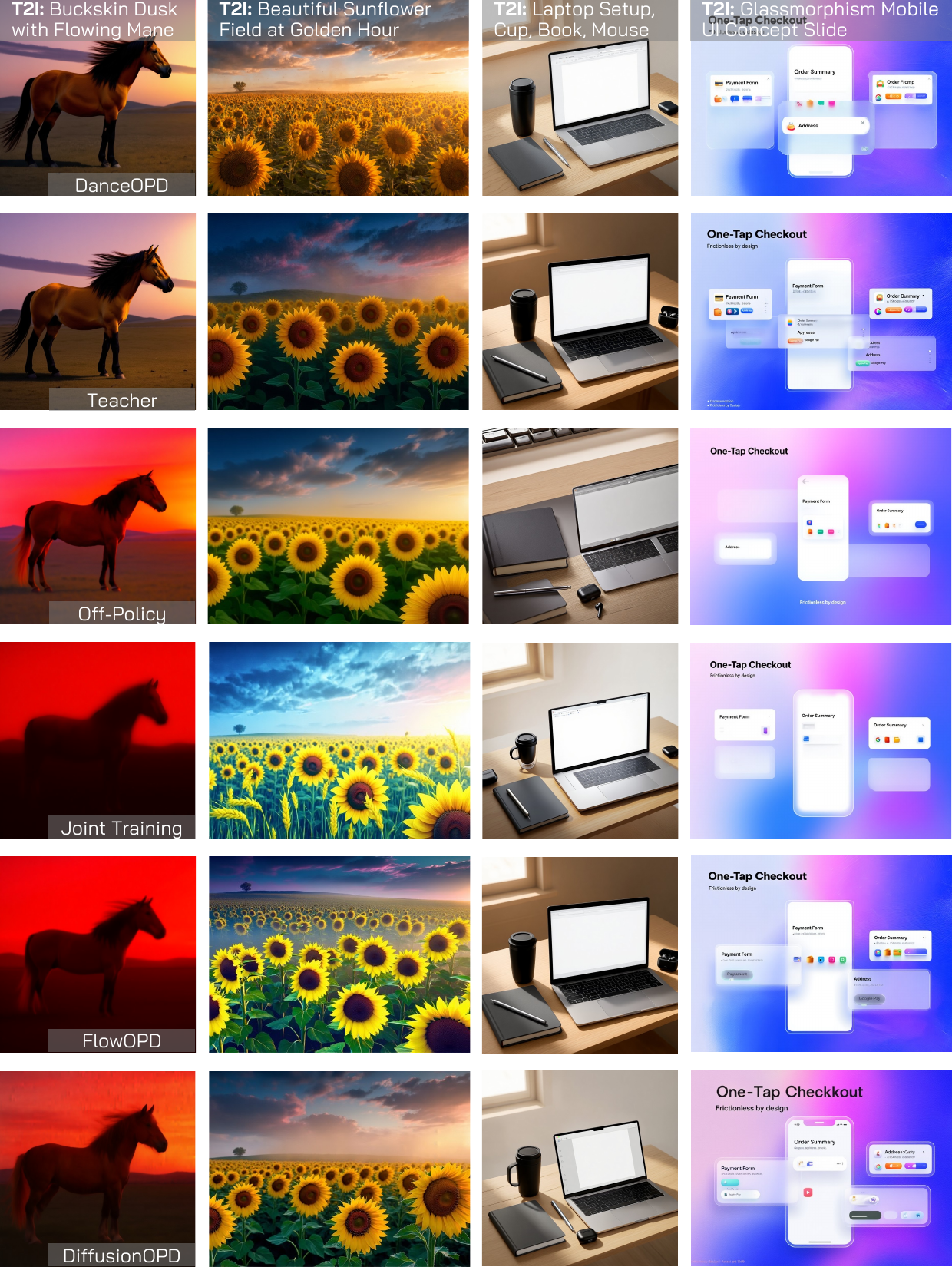}
    \vspace{-0.5cm}
    \caption{\textbf{T2I and Edit Composition: Text-to-Image Ability.} DanceOPD retains strong text-to-image generation quality while learning editing capabilities, whereas several baselines introduce color shift, blur, or layout degradation.}
    \label{fig:t2i_1}
\end{figure}

\noindent\textbf{A. T2I and Edit Composition.}
We first test whether \name{} can add general editing ability to a T2I student while preserving its base generation capability. DanceOPD improves the GEditBench average over the best reproduced OPD baseline by $8.1\%$ and over the edit source by $8.5\%$, while improving GenEval overall over the T2I source by $2.0\%$ and over the strongest composition baseline by $1.6\%$. The gains are especially clear on edit categories that require larger visual changes. Compared with DiffusionOPD, DanceOPD improves background change by $21.9\%$, style change by $21.3\%$, and color alteration by $5.5\%$. These results indicate that routed on-policy field matching can integrate the edit field into the student without eroding the base generation field. The qualitative examples show the same pattern, where DanceOPD follows scene, style, material, and object-level edit instructions while retaining prompt-following T2I generation. Fig.~\ref{fig:viz}, Fig.~\ref{fig:editing_1}, Fig.~\ref{fig:editing_2}, Fig.~\ref{fig:editing_3}, Fig.~\ref{fig:t2i_1}, and Fig.~\ref{fig:shared-object-edits} show the same pattern for T2I and Edit Composition: DanceOPD follows scene, style, material, and object-level edit instructions while retaining prompt-following T2I generation.

\noindent\textbf{B. Local and Global Edit Composition.}
We then evaluate whether one student can compose local editing, which emphasizes preservation, with global editing, which requires broader visual transformation. \name{} improves the GEditBench average over the best competing composition baseline by $16.1\%$ and over the local edit source by $7.9\%$, while improving GenEval overall over the strongest composition baseline by $2.5\%$. The improvement is concentrated in categories that require global or attribute-level changes. Compared with the best competing composition baseline in each category, DanceOPD improves background change by $33.5\%$, style change by $12.9\%$, and color alteration by $11.6\%$. This supports the claim that capability-field composition is better handled by routed field queries than by parameter averaging or mixed supervision.

\begin{table*}[t]
  \centering
  \vspace{-0.2cm}
  \setlength{\tabcolsep}{2pt}
  \caption{\textbf{Main Multi-Capability Composition Results with Two Different Settings.} We compare \name{} with Joint Training, weight merging, off-policy distillation, and other OPD baselines on GEditBench-EN~\cite{liu2025step1x} for image editing and GenEval~\cite{ghosh2023geneval} for text-to-image. DanceOPD achieves better capability composition.}
  \vspace{-0.2cm}
  \label{tab:zimage-multi}
  \resizebox{\textwidth}{!}{%
  \begin{tabular}{lccccccc|ccccccc}
    \toprule
    \multirow{2}{*}{\textbf{Method}} & \multicolumn{7}{c|}{\textbf{GEditBench-EN}~\cite{liu2025step1x}} & \multicolumn{7}{c}{\textbf{GenEval}~\cite{ghosh2023geneval}}\\
    & \textbf{Subj-Add} & \textbf{Subj-Rep} & \textbf{Bg-Chg} & \textbf{Style-Chg} & \textbf{Color-Alt} & \textbf{Subj-Rem} & \textbf{Avg}
    & \textbf{Sing} & \textbf{Two} & \textbf{Cnt} & \textbf{Col} & \textbf{Pos} & \textbf{C-Attr} & \textbf{Overall}\\
    \midrule
    \rowcolor{gray!10}
    \multicolumn{15}{l}{\textbf{Base Model}}\\
    T2I & -- & -- & -- & -- & -- & -- & -- & 0.950 & 0.939 & 0.938 & 0.947 & 0.520 & 0.700 & 0.832\\
    Edit & 6.033 & 5.417 & 4.490 & 3.923 & 4.889  & 4.828 & 4.930 & 0.838 & 0.828 & 0.713 & 0.840 & 0.580  & 0.470 & 0.711\\
    Local Edit & 5.555 & 5.742 & 4.856 & 3.817 & 4.581 & 6.017 & 5.095 & 0.988 & 0.929 & 0.813 & 0.862 & 0.600 & 0.570 & 0.793\\
    Global Edit & 3.119 & 4.414 & 4.040 & 5.209 & 4.287 & 1.433 & 3.750 & 0.950 & 0.939 & 0.838 & 0.872 & 0.600 & 0.650 & 0.808\\
    \midrule
    \rowcolor{gray!10}
    \multicolumn{15}{l}{\textbf{A. T2I and Edit Composition}}\\
    Joint Training & 5.386 & 5.627 & 4.283 & 3.688 & 3.624 & 5.093 & 4.617 & 0.975 & 0.929 & 0.963 & 0.894 & 0.540 & 0.550 & 0.808\\
    Weight Merge & -- & -- & -- & -- & -- & -- & -- & 1.000 & 0.929 &0.925 &0.894 &0.610 & 0.670 &  0.836\\
    Off-Policy Distill. & 4.882& 5.026& 4.289& 4.497& 4.679& 3.797& 4.528& 0.975& 0.919& 0.887& 0.894& 0.580& 0.650& 0.818\\
    DiffusionOPD~\cite{li2026diffusionopd} & 5.488& 5.850& 4.242& 4.303& 4.588& 5.211& 4.947& 1.000& 0.929& 0.975& 0.894& 0.590& 0.610& 0.833 \\
    Flow-OPD~\cite{fang2026flow} & 6.014& 5.214& 4.467& 3.957& 4.793& 4.681& 4.854& 0.975& 0.909& 0.888& 0.926& 0.550& 0.640& 0.814  \\
    \rowcolor{vefblue3!9}
    \textbf{\ding{72}\,\name{} (Ours)} &   5.681 & 5.857 & 5.173 &  5.218 &  4.840 & 5.310 &5.347 & 0.988& 0.939& 0.963& 0.894& 0.640& 0.670& 0.849 \\
    \midrule
    \rowcolor{gray!10}
    \multicolumn{15}{l}{\textbf{B. Local and Global Edit Composition}}\\
    Joint Training & 4.632 & 5.393 & 4.128 & 3.941 & 4.093 & 5.086 & 4.546 & 0.975 & 0.939 & 0.913 & 0.851 & 0.600 & 0.650 & 0.821\\
    Weight Merge & 4.434 & 4.776 & 4.380 & 5.263 & 5.206 & 4.229 & 4.715 & 0.988& 0.909& 0.875& 0.904& 0.600& 0.590& 0.811\\
    Off-Policy Distill. & 5.008& 4.683& 4.543& 4.772& 5.075& 4.336& 4.736& 0.975& 0.889& 0.900& 0.872& 0.570& 0.580& 0.798\\
    DiffusionOPD~\cite{li2026diffusionopd} & 4.704& 5.310& 4.502& 4.012& 4.977& 4.462& 4.661& 0.975& 0.899& 0.925& 0.883& 0.620& 0.630& 0.822 \\
    Flow-OPD~\cite{fang2026flow} & 4.524& 4.647& 4.610& 5.232& 5.037& 4.025& 4.679 & 0.975& 0.949& 0.875& 0.872& 0.630& 0.660& 0.827  \\
    \rowcolor{vefblue3!9}
    \textbf{\ding{72}\,\name{} (Ours)} & 5.178 &  5.549 &6.153 &   5.944 &   5.812 &  4.348&5.498 & 1.000& 0.949& 0.925& 0.926& 0.650& 0.640& 0.848 \\
    \bottomrule
  \end{tabular}
  }
  \vspace{0.2cm}
\end{table*}

\begin{figure*}[t]
    \centering
    \includegraphics[width=\textwidth]{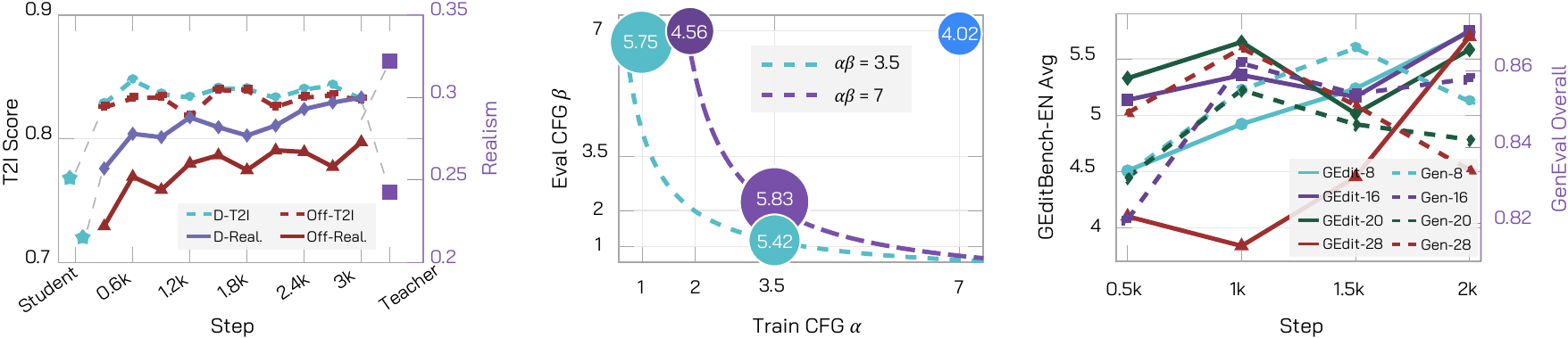}
    \vspace{-0.59cm}
    \caption{\textbf{Field Absorption and Rollout Diagnostics.} (a) DanceOPD absorbs a realism field more effectively than off-policy distillation while preserving the base text-to-image capability. (b) Training-time and inference-time CFG scales compose, where $\alpha\beta$ denotes the effective CFG in practice. DanceOPD absorbs the training CFG.  (c) Moderate rollout discretization suffices for stable field queries during training.}
    \label{fig:cfg-absorption-map}
\end{figure*}

\begin{figure}[t]
    \centering
    \includegraphics[width=\textwidth]{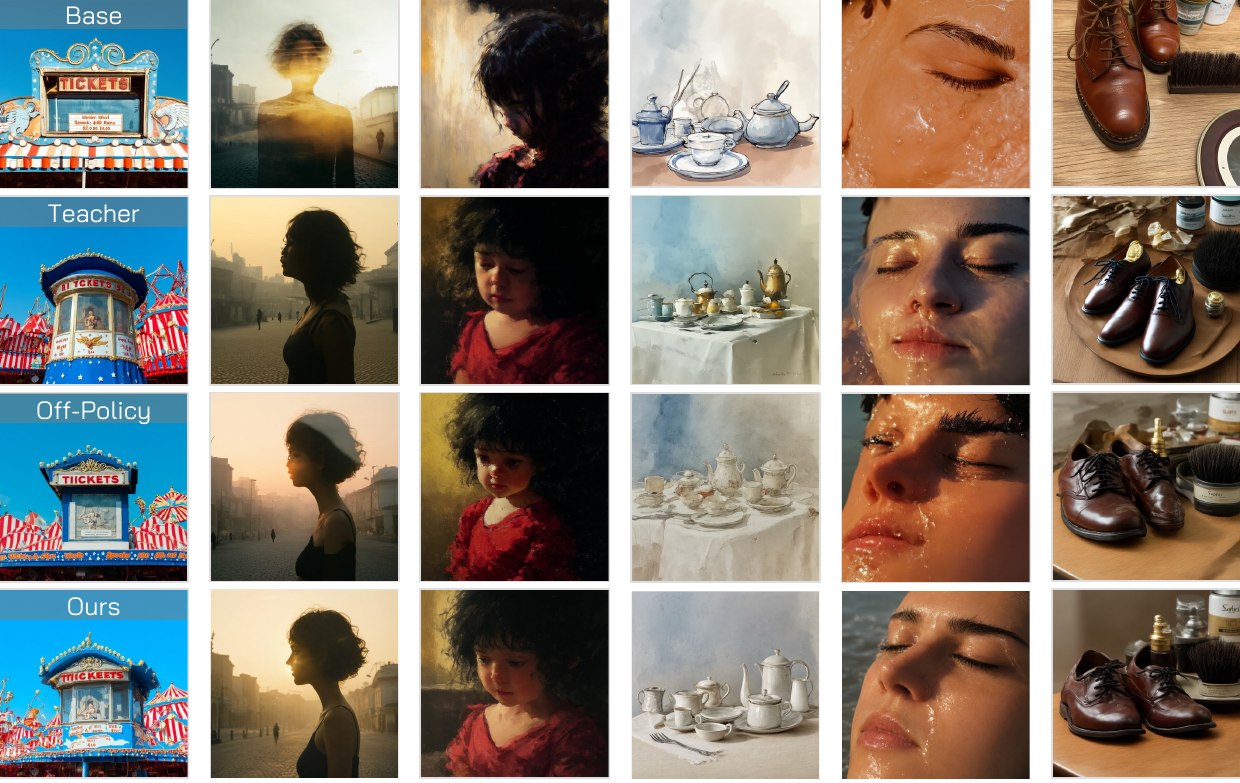}
    \vspace{-0.5cm}
    \caption{\textbf{Realism-Field Absorption.} DanceOPD absorbs the teacher's realism-oriented field more effectively than off-policy distillation, shifting samples toward more photorealistic texture, lighting, and visual statistics while preserving the base model's prompt-following and content generation ability.}
    \label{fig:absord}
\end{figure}

\noindent\textbf{C. Realism Absorption.}
We next evaluate whether \name{} can absorb a realism-oriented quality field that shifts generations toward more photorealistic texture, lighting, and visual statistics while preserving the original T2I capability. DanceOPD improves the realism reward over off-policy distillation by $9.9\%$ and closes $85.3\%$ of the student-to-teacher reward gap. At the same time, its T2I score improves over the student anchor by $7.6\%$ and matches off-policy distillation within $0.1\%$, indicating that the student moves toward the realism teacher without sacrificing base generation behavior. The qualitative results in Fig.~\ref{fig:absord} show the same trend. In the ticket-booth example, our model enhances lighting contrast, structural details, and photographic sharpness while keeping the original scene layout. In the portrait and face examples, it improves skin texture, illumination, and water-reflection details without changing the main subject. In the tea-set and leather-shoe examples, DanceOPD produces more coherent material appearance and realistic surface reflections, whereas off-policy distillation is more prone to washed-out structure or over-saturated texture. These results suggest that on-policy querying is useful not only for task composition, but also for absorbing style- or quality-oriented fields while preserving prompt content.

\noindent\textbf{D. CFG Absorption.}
We finally evaluate whether \name{} can absorb an operator-defined classifier-free guidance field into the student, so that part of the guidance effect is internalized rather than applied only at inference time~\cite{ho2022classifier}. We instantiate this case by treating a CFG-guided velocity field as the frozen target field and training the student to approximate it with a single forward pass. This setting also exposes an evaluation caveat: absorbed guidance and external inference-time guidance are not independent. If the absorbed target uses guidance scale $\alpha$ and inference applies scale $\beta$ again, the effective guidance strength is approximately $\alpha\beta$, so evaluating the absorbed student under the default external-CFG convention can over-guide the model. Fig.~\ref{fig:cfg-absorption-map} (b) confirms this interaction. The best measured composition improves the GEditBench average over train-only absorption by $7.6\%$ and over eval-only CFG by $1.4\%$. The gains over train-only absorption are especially visible in subject replacement, background change, color alteration, and subject removal, with relative improvements of $10.3\%$, $22.5\%$, $6.6\%$, and $11.2\%$, respectively. By contrast, excessive composition of absorbed and external guidance reduces the score by $31.2\%$ relative to the best measured composition.

\subsection{Multi-Teacher Extension}
\label{sec:multi-teacher-ext}

After the main composition results, we study how several capability fields should supervise the same student during one training process. This setting is related to multi-teacher distillation and task-customized knowledge amalgamation~\cite{hinton2015distilling,shen2019customizing}, but the key issue here is the field query itself. When several fields enter the same update, the student can lose sample-level semantic identity through soft teacher mixing, or overuse correlated states through dense rollout supervision. We therefore use this subsection to isolate target-field ambiguity and trajectory-query correlation, the two query-level failure modes introduced in Sec.~\ref{sec:hard-routing} and Sec.~\ref{sec:query-design}.

The diagnostics vary three existing choices while keeping the same capability fields. The first comparison tests hard routing against soft all-teacher mixing. The second tests step alternation against same-step accumulation, where $G$ denotes the number of capability buckets accumulated before one optimizer step. The third varies dense trajectory supervision, where $K$ denotes the number of queried states per rollout, and uses SDE rollout only as a diagnostic control. Table~\ref{tab:multi-teacher-aggregation} reports the full per-category values.

\noindent\textbf{Hard Routing versus Soft Teacher Mixing.}
Hard routing is needed because each training sample should be supervised by the capability field that matches its semantic role. Soft all-teacher mixing removes this sample-level identity by averaging all teacher fields into one target, so the target no longer cleanly represents the intended T2I, local edit, or global edit query. This difference is consistent across objectives. With MSE, hard routing improves the average over soft mixing by $15.2\%$, with gains on subject addition, background change, style change, color alteration, and subject removal. The largest gains appear on background change and subject removal, where hard routing improves by $20.8\%$ and $26.8\%$, respectively. With KL-$\bar{\sigma}^2$, hard routing still improves the average over soft mixing by $10.6\%$, including $14.5\%$ on style change and $14.4\%$ on subject removal. These results show that the main issue is target-field construction, rather than only the choice of matching objective.

\begin{figure*}[t]
  \centering
  \includegraphics[width=\textwidth]{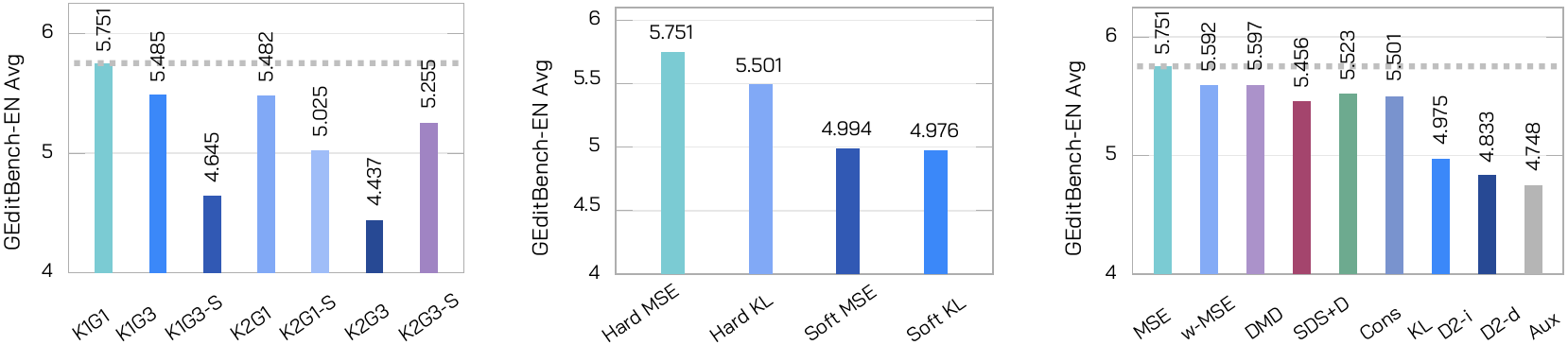}
  \vspace{-0.5cm}
  \caption{\textbf{Routing, Objective, and Dense-Query Diagnostics.} Hard routing with single-query MSE is the strongest default; same-step accumulation, soft teacher mixing, and dense correlated queries degrade performance.}
  \label{fig:routing-dense-diagnostics}
\end{figure*}

\noindent\textbf{Same Step Multi-Teacher Accumulation.}
Step alternation keeps each optimizer update tied to one routed capability bucket. Same-step accumulation keeps the routed target for each individual sample, but combines several capability buckets before one optimizer step. This tests whether routing remains sufficient when multiple capability fields affect the same update. Compared with the single-query step-alternation default, same-step accumulation with $K{=}1$ and $G{=}3$ loses $4.6\%$ on average. The change is not a uniform degradation across all categories. Style change and color alteration improve, but subject addition and subject removal drop by $17.5\%$ and $13.5\%$, showing that same-step accumulation shifts the balance between capabilities rather than preserving all of them. When dense supervision is also added, the $K{=}2,G{=}3$ case loses $22.8\%$ on average, with subject addition and subject removal dropping by $28.9\%$ and $46.0\%$. This shows that capability conflict can reappear at the optimizer-update level, even when every individual sample keeps a valid routed target~\cite{yu2020gradient,liu2021conflict}.

\noindent\textbf{Dense Query Drift Control.}
The dense-query diagnostic tests whether the poor dense same-step result is tied to trajectory-query correlation. In the $K{=}2,G{=}3$ stress case, replacing ODE rollout with SDE rollout improves the average by $18.4\%$. The recovery is strongest on subject removal and subject addition, which improve by $62.0\%$ and $29.0\%$, and it also improves subject replacement and background change by $15.4\%$ and $10.8\%$. However, the rescued result still remains $8.6\%$ below the single-query default, and it is weaker on style change, color alteration, and subject removal by $18.7\%$, $9.8\%$, and $12.5\%$. SDE rollout also hurts the two control settings by $15.3\%$ and $8.3\%$. Thus, stochastic rollout noise is useful as diagnostic evidence for trajectory-query correlation, but it is not a better default. The safer design is to avoid correlated dense supervision in the first place and use a single semantic-side query.

\subsection{Ablation Study}
\label{sec:exp-ablation}

\begin{figure*}[t]
    \centering
    \includegraphics[width=\textwidth]{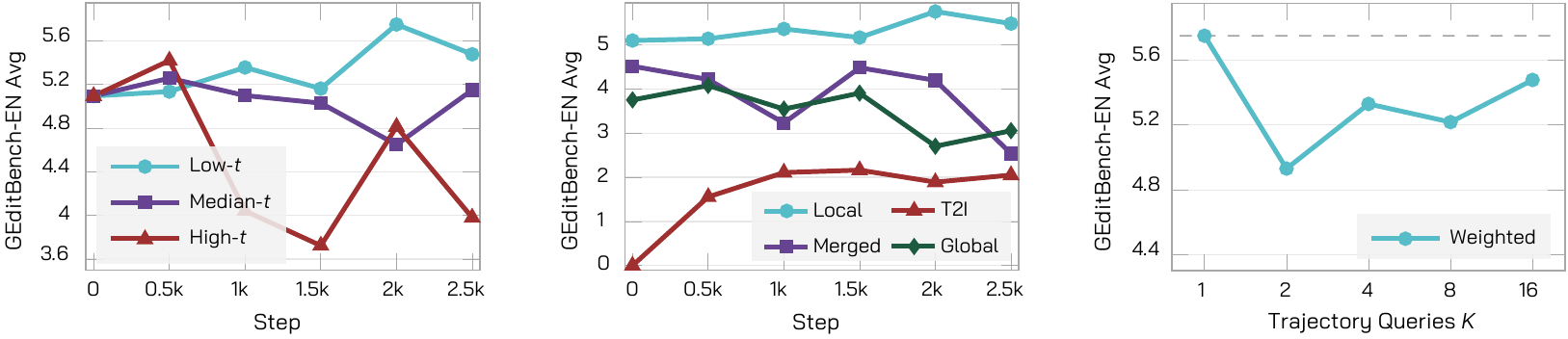}
    \vspace{-0.62cm}
    \caption{\textbf{Ablation Trends.} Semantic-side low-$t$ queries and local-edit initialization give the most reliable gains, while increasing the number of trajectory queries does not improve performance.}
    \label{fig:timestep-init-ablation-curves}
\end{figure*}

The previous subsection studied how multiple capability fields should enter the same student update. We now ablate the remaining training choices of \name{} under the same frozen capability fields and the same evaluation protocol. These choices include the number of rollout steps used to obtain student states, the timestep region where the teacher field is queried, the number of queried states from one rollout, the local matching objective, and the student initialization. Fig.~\ref{fig:timestep-init-ablation-curves} summarizes the main trends, while Table~\ref{tab:ablation-summary} and Table~\ref{tab:rollout-step-sensitivity} report the detailed values.

\noindent\textbf{Rollout Steps.}
The rollout is used to obtain student-visited states for teacher queries, which generally need to be consistent with inference.
However, we found that in DanceOPD, the training rollout length does not need to match the evaluation sampler exactly. This may be because our solver is ODE~\cite{song2020denoising,lu2022dpm,karras2022elucidating}. In the DanceOPD experiments, at $2000$ steps, a moderate rollout length is sufficient. At $2$k steps, the $16$-step rollout gives the strongest GEditBench average, improving over the 8-step, $20$-step, and $28$-step variants by $0.2\%$, $3.0\%$, and $0.9\%$, respectively. It also improves GenEval overall over these variants by $0.7\%$, $1.9\%$, and $2.9\%$. Although the 28-step rollout is competitive on some edit sub-scores, the $16$-step rollout improves subject removal over it by $33.7\%$ and gives better T2I preservation. This shows that longer rollout discretization does not automatically provide better supervision for field matching.

\noindent\textbf{Timestep Query.}
Capability-specific supervision is most useful on the semantic, low-noise side of the trajectory, consistent with timestep-sensitive image editing and distillation analyses~\cite{wang2023not,chen2024tino,lin2024schedule}. At $2$k steps, low-$t$ querying improves the GEditBench average over median-$t$ and high-$t$ querying by $23.7\%$ and $19.5\%$. Compared with median-$t$, low-$t$ improves subject addition by $35.9\%$, background change by $36.1\%$, color alteration by $14.0\%$, and subject removal by $42.3\%$. Compared with high-$t$, low-$t$ improves subject addition by $46.1\%$ and style change by $57.1\%$. These results support the design choice of querying the teacher field on the semantic side of the student rollout.

\noindent\textbf{Number of Trajectory Queries.}
Using more states from the same rollout does not necessarily improve supervision, because states along one rollout are highly correlated, echoing recent observations that rollout position and rollout length affect OPD stability~\cite{ziheng2026moreearlystoppingrollout}. The single-query default outperforms the weighted dense-query variants across $K{=}2,4,8,16$, improving the GEditBench average by $16.6\%$, $7.9\%$, $10.2\%$, and $12.2\%$, respectively. Even the strongest weighted dense-query variant remains below the single-query default. This confirms that one semantic-side query per rollout is not only more efficient, but also more reliable for matching the routed teacher field.

\noindent\textbf{Objective Design.}
Plain velocity MSE gives the strongest average among the tested local matching objectives. It improves over timestep-weighted MSE and DMD-EMA hybrid by $2.8\%$, over consistency matching by $4.1\%$, and over KL-$\bar{\sigma}^2$ matching by $4.5\%$. The gap is larger for DMD2 and AuxFeat variants, where plain velocity MSE improves the average by $15.6\%$ to $21.1\%$. Some alternatives improve individual sub-scores, but they do not give the same overall balance across edit categories. Since the target is a routed velocity field queried at a fixed student state, direct velocity regression gives the most stable update.

\noindent\textbf{Initialization.}
Initialization matters because the student rollout determines where teacher fields are queried early in training. The best default is the checkpoint with the strongest relevant capability, rather than merged initialization or an unrelated anchor. At $2$k steps, local edit initialization improves the GEditBench average over merged initialization by $37.2\%$, over global edit initialization by $112.8\%$, and over T2I initialization by $204.4\%$. Compared with merged initialization, local edit initialization improves subject addition by $48.8\%$, subject replacement by $32.1\%$, background change by $28.7\%$, color alteration by $29.2\%$, and subject removal by $124.0\%$. This indicates that a stronger initial student rollout makes field matching more effective.

\begin{figure}[H]
    \centering
    \vspace{-1cm}
    \includegraphics[width=\textwidth]{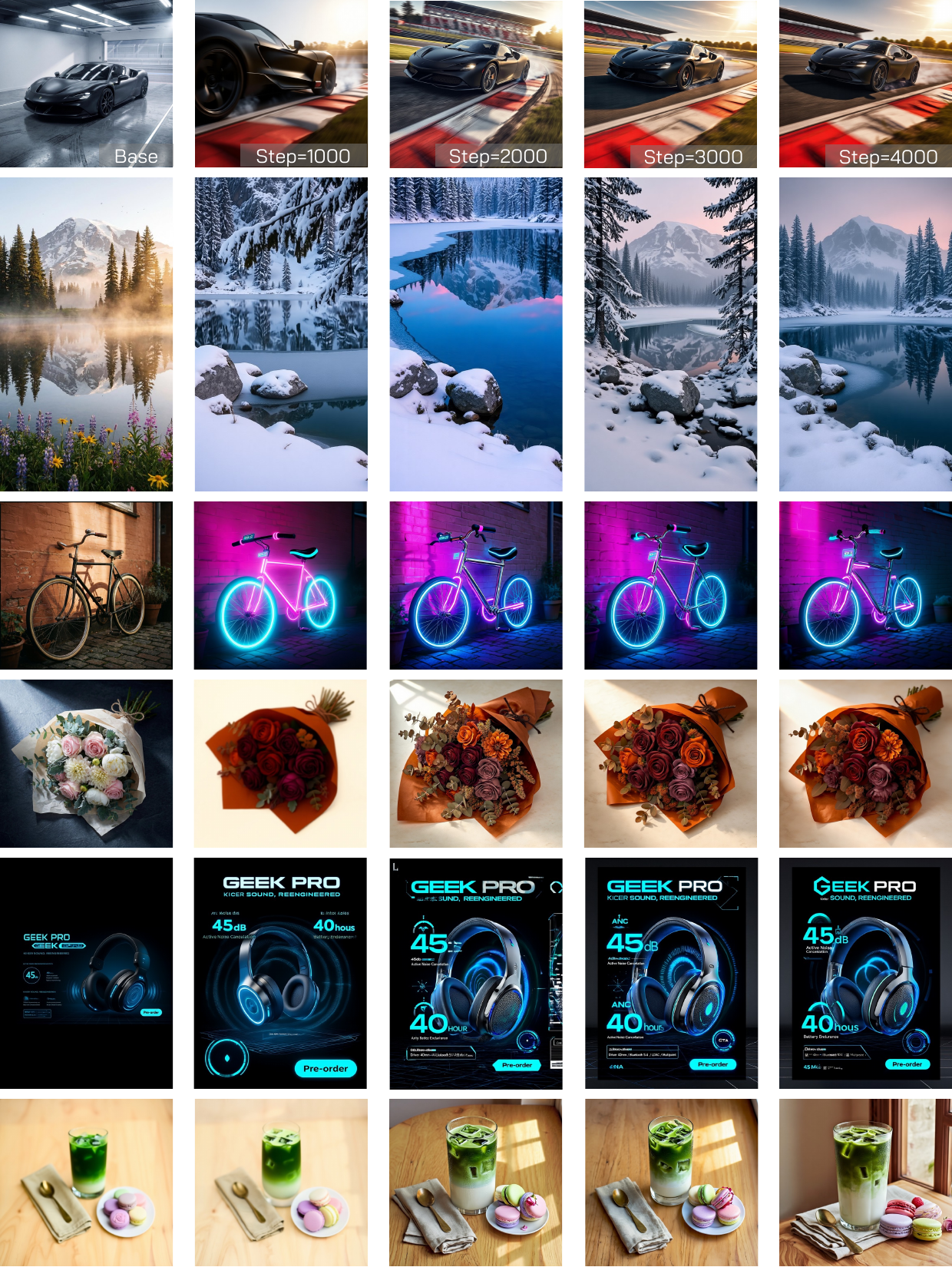}
    \vspace{-0.5cm}
    \caption{\textbf{Training Progression in Local and Global Edit Composition.} As distillation proceeds, \name{} progressively absorbs the target edit capability while maintaining scene identity and prompt alignment.}
    \label{fig:m_stps}
\end{figure}

\subsection{Capabilities Emergence}
\label{sec:capabilities-emergence}

\begin{figure*}[t]
    \centering
    \includegraphics[width=\textwidth]{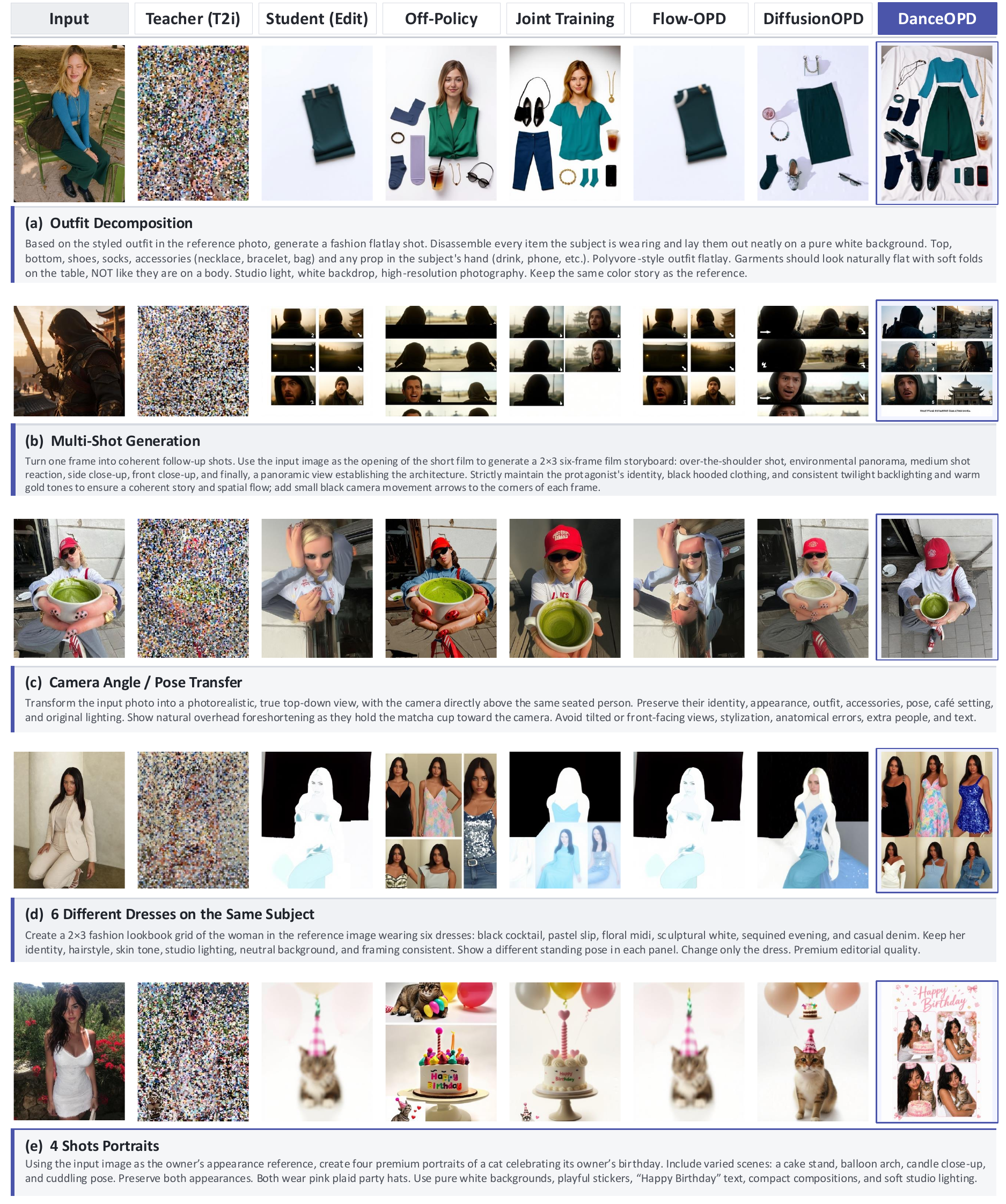}
    \vspace{-0.2cm}
    \caption{\textbf{Zero-Shot Capability Emergence on Unseen Reference-Conditioned Tasks.}
    Neither the T2I teacher nor the edit-initialized student is explicitly trained for these tasks. DanceOPD nevertheless exhibits coherent outfit decomposition, identity-preserving multi-shot generation, camera-angle and pose transfer, and reference-based visual recomposition, while the competing methods all fail.}
    \label{fig:ood-capability-emergence}
\end{figure*}

An unexpected qualitative finding arises when we evaluate zero-shot tasks that are absent from the training distribution. Using the final checkpoint from the T2I and Edit Composition setting in Table~\ref{tab:zimage-multi}, we test reference-conditioned generation tasks for which neither source model was explicitly trained. As shown in Fig.~\ref{fig:ood-capability-emergence}, DanceOPD is the only evaluated method that consistently exhibits capabilities unavailable to both the T2I teacher and the edit-initialized student. The T2I teacher is the text-to-image Z-Image model, which cannot interpret the additional input image and therefore produces Gaussian-noise outputs under the editing interface. The student is initialized from a Z-Image Edit model trained on OmniEdit. However, OmniEdit contains only local and global editing data and does not include the novel reference-generation tasks considered here, such as multi-shot generation, viewpoint changes, or outfit decomposition.

In outfit decomposition [Fig.~\ref{fig:ood-capability-emergence}(a)], DanceOPD separates the clothing into semantically meaningful individual garments rather than copying the complete input. In multi-shot generation [Fig.~\ref{fig:ood-capability-emergence}(b)], it preserves the identity of the subject while producing subsequent views with clear changes in shot and viewpoint. Flow-OPD, by contrast, remains substantially closer to the structure of the source image. This behavior indicates that DanceOPD does not merely select or reproduce the output of one teacher. Because the distinct capability fields remain well separated within a shared student, the model can recombine existing behaviors and exhibit stronger capabilities when a test request lies between the original training tasks.

This result is also notable from the perspective of timestep specialization. Under the conventional coarse-to-fine interpretation of diffusion trajectories, high-noise, high-$t$ states are primarily responsible for establishing global composition and spatial layout, whereas low-noise, low-$t$ states mainly refine local appearance and details and are therefore not expected to reconstruct a substantially different layout~\cite{wang2023not}. DanceOPD, however, is supervised through only one low-$t$ query, yet its outputs on these held-out tasks differ qualitatively from those of both the T2I teacher and the edit-initialized student. The model therefore does more than copy the layout of either endpoint; it appears to synthesize a new spatial organization from the composed capability fields. This departure from the usual timestep interpretation provides another intriguing indication of emergent behavior.

Flow-OPD and DiffusionOPD instead tend to copy the behavior associated with the editing branch. We hypothesize that the additional image input makes these requests more likely to be treated as editing tasks, causing their outputs to remain closer to the edit-initialized student, while Off-Policy Distillation and Joint Training show inconsistent behavior. We further conjecture that the stronger zero-shot generalization of DanceOPD is related to its single-query design. By avoiding dense, correlated supervision along the same rollout, single-query field matching may preserve more clearly separated and recomposable capability fields, enabling novel behaviors to emerge on previously unseen tasks.
\section{Conclusion}
We introduced \name{}, an on-policy generative field distillation framework for composing heterogeneous generative capabilities into a single flow-matching student. The key view is to treat each frozen capability source as a velocity field over a shared generative state space and to formulate capability composition as a field-query problem. This perspective exposes three query-induced alignment challenges, including target-field ambiguity, state-distribution mismatch, and trajectory-query correlation. Correspondingly, our model uses hard-routed sample-wise field matching to preserve the semantic identity of each capability, queries the routed field on student-visited states from the current rollout, and uses one semantic-side low-noise query per sample to avoid correlated within-trajectory supervision. With a plain velocity MSE objective, this design composes and absorbs capability fields without relying on joint training, parameter-space merging, or external score composition. 

Experiments on T2I and editing composition, local and global editing composition, realism-field absorption, and CFG absorption show that DanceOPD strengthens the target capability while preserving the anchor capability. Ablations further support the role of hard routing, student-induced query states, semantic-side single-query supervision, and direct velocity regression. These results suggest that on-policy generative field distillation is a practical route toward scalable multi-capability visual generation.

\vspace{0.2cm}

\cleardoublepage

\section{Limitations and Discussions}
\label{sec:limitations}

\noindent\textbf{Shared Field Support.}
The current formulation assumes that frozen capability sources expose compatible velocity fields over a shared generative state space. In our experiments, this assumption is satisfied because the sources are built from the same backbone family, latent representation, scheduler convention, and velocity parameterization. This requirement is not unique to DanceOPD: LLM OPD methods usually require teacher and student distributions to be defined over the same token space before token-level KL or asymmetric KL can be applied~\cite{gu2024minillm,agarwal2024policy,jia2026asymmetric}, and flow-matching OPD methods also require compatible transition, score, or velocity parameterizations for teacher and student matching~\cite{li2026diffusionopd,jiang2026d,fang2026flow}.

\noindent\textbf{Predefined Routing.}
Our implementation uses predefined capability buckets and sample-wise hard routing. This is a standard and stable OPD-style design when the supervision source is known from the task or data identity, and it is well matched to our setting, where T2I, local editing, global editing, style, and guidance-field examples are easy to separate. However, this assumption becomes weaker when task boundaries are ambiguous or when a prompt requires several capabilities at once. A natural extension is to introduce a verifier or reward model that assigns routes according to model behavior or predicted edit success, connecting our field-routed distillation to reward- and verifier-based evaluation or supervision signals~\cite{kirstain2023pick,wu2023human,xu2023imagereward,ku2024viescore,fang2026rubric}.

\vspace{0.2cm}
\section{Theoretical Details}
\label{app:theory}

This appendix provides the theoretical motivation behind \name{} and follows the same field-query logic as the main text. Given frozen capability fields over a shared flow state space, the student must decide which field to query, which student-induced state to query, and how many states from the same rollout to use. 

The local analyses below explain why plain velocity MSE is a natural field-matching objective, why teacher fields should be queried on stop-gradient states from the current student rollout, why mixing several fields inside one sample target creates target-field ambiguity, and why dense same-rollout queries can suffer from trajectory-query correlation. The experiments then test these predicted failure modes and design choices in full image-generation models.

\vspace{0.2cm}

\subsection{KL-MSE Equivalence for Velocity-Field Distillation}
\label{app:kl-mse}

DanceOPD uses plain velocity MSE in the main method. Here, we show why this objective is the natural form of a KL-style distillation loss when both the student and frozen capability source define local Gaussian transition kernels over the same flow state space.

Consider a small reverse-time step from $t$ to $t-\Delta t$. Let the student and the routed capability field induce local Gaussian kernels with shared covariance $\sigma_t^2 I$:
\begin{align}
  p_\theta(z_{t-\Delta t}\mid z_t,c)
  &= \mathcal{N}\bigl(z_t - \Delta t\,v_\theta(z_t,t,c),\,\sigma_t^2 I\bigr),\\
  p_m(z_{t-\Delta t}\mid z_t,c)
  &= \mathcal{N}\bigl(z_t - \Delta t\,v_m(z_t,t,c),\,\sigma_t^2 I\bigr).
\end{align}
For two Gaussians with identical covariance, the forward KL has a closed form:
\begin{equation}
  D_{\mathrm{KL}}(p_m\,\|\,p_\theta)
  =
  \frac{\Delta t^2}{2\sigma_t^2}
  \left\|v_\theta(z_t,t,c)-v_m(z_t,t,c)\right\|_2^2.
  \label{eq:app-kl-mse}
\end{equation}
The reverse KL has the same quadratic form under the same covariance assumption. Thus, KL-style local transition matching reduces to a timestep-weighted velocity MSE. This is consistent with flow matching, which directly regresses vector fields along a probability path~\cite{lipman2022flow}, and with diffusion and SDE formulations where Gaussian perturbation kernels yield quadratic score or velocity matching objectives~\cite{sohl2015deep,song2019generative,ho2020denoising,nichol2021improved,song2020denoising,dhariwal2021diffusion,song2020score}. In the main method, we use the unweighted MSE because the teacher target is deterministic, and our ablations find the plain objective more stable than the tested weighting variants.

\vspace{0.2cm}

\subsection{On-Policy Query Distribution and Stop-Gradient Rollout}
\label{app:onpolicy-sg}

Let $\Phi_\theta^t$ denote the numerical solver that maps an initial noise $z_T$ to an intermediate state at time $t$ using the current student velocity, and let $p_T$ denote the initial-noise distribution. DanceOPD samples a semantic-side coordinate and maps it to a physical timestep before querying the rollout:
\begin{equation}
  s\sim q_{\mathrm{sem}}(s),\quad t=t(s),\quad z_t^\theta=\Phi_\theta^t(z_T,c), \qquad z_T\sim p_T.
  \label{eq:app-onpolicy-query}
\end{equation}
Conditioned on a routed sample $m\sim\pi$ and $(x,c)\sim\mathcal{D}_m$, the implemented loss uses the sampled state as a query point, that is:
\begin{equation}
  \mathcal{L}_{\mathrm{sg}}
  =
  \mathbb{E}_{m\sim\pi,\,(x,c)\sim\mathcal{D}_m,\,z_T\sim p_T,\,s\sim q_{\mathrm{sem}}}\left[
  \left\|
  v_\theta(\mathrm{sg}(z_t^\theta),t,c)
  -v_m(\mathrm{sg}(z_t^\theta),t,c)
  \right\|_2^2
  \right],\qquad t=t(s),
  \label{eq:app-sg-loss}
\end{equation}
where $\mathrm{sg}(\cdot)$ denotes stop-gradient. Thus, the gradient is:
\begin{equation}
  \nabla_\theta\mathcal{L}_{\mathrm{sg}}
  =
  2\,\mathbb{E}_{m,(x,c),z_T,s}\left[
  \bigl(v_\theta-v_m\bigr)^\top
  \nabla_\theta v_\theta(\mathrm{sg}(z_t^\theta),t,c)
  \right],
  \label{eq:app-sg-grad}
\end{equation}
where $v_\theta$ and $v_m$ inside the inner product are evaluated at $(\mathrm{sg}(z_t^\theta),t,c)$, with the capability target detached. This is the practical first-order estimator: it avoids differentiating through every solver step while still querying the field on the student's current state distribution. This corresponds to the second query choice in the main text, where the teacher field is evaluated on the states visited by the current student rather than on fixed off-policy states.

The reason to query on-policy can be stated as a simple mismatch bound. Suppose an off-policy method queries a state $\tilde z_t$ while the deployed student visits $z_t^\theta$. If the routed capability field is $L_m$-Lipschitz in $z$, then
\begin{equation}
  \left\|v_m(z_t^\theta,t,c)-v_m(\tilde z_t,t,c)\right\|_2
  \le L_m\left\|z_t^\theta-\tilde z_t\right\|_2.
  \label{eq:app-offpolicy-bound}
\end{equation}
Therefore, teacher supervision collected far from the student rollout can be a biased local target for deployment-time states. This mirrors the covariate-shift motivation behind dataset aggregation in imitation learning~\cite{ross2011reduction}, but here the states are continuous diffusion and flow trajectories rather than discrete actions.

\vspace{0.2cm}

\subsection{Smoothness of Capability Fields on Student-Rolled States}
\label{app:smoothness}

A natural concern is that a student rollout may visit states different from those produced by a frozen capability source. The following standard flow-matching view explains why this does not automatically invalidate the queried target.

For a rectified-flow interpolant with data endpoint $x_0$ and noise endpoint $x_1$, the intermediate state can be written as:
\begin{equation}
  z_t=(1-t)x_0+t x_1.
\end{equation}
The optimal flow-matching velocity is a conditional expectation~\cite{lipman2022flow}, that is:
\begin{equation}
  v^\star(z,t)=\mathbb{E}\!\left[x_1-x_0 \mid z_t=z\right].
  \label{eq:app-fm-condexp}
\end{equation}
Conditional expectations over Gaussian-convolved marginals vary smoothly with $z$ away from degenerate endpoints. Let $v_m$ be a trained capability field whose approximation error to its ideal field is bounded by $\delta_m$ on the relevant marginal support, and let the ideal field be $L_t$-Lipschitz in $z$. For a student-visited state $z_t^\theta$ and a nearby reference state $\tilde z_t$ satisfying $\|z_t^\theta-\tilde z_t\|_2\le\varepsilon$, we have:
\begin{equation}
  \left\|v_m(z_t^\theta,t,c)-v_m(\tilde z_t,t,c)\right\|_2
  \le L_t\varepsilon+2\delta_m.
  \label{eq:app-teacher-robust}
\end{equation}
The bound follows by adding and subtracting the ideal field and applying the triangle inequality. Its role is not to prove that arbitrary off-policy states are safe; rather, it formalizes the regime used by \name{}: when the current student remains within a bounded neighborhood of the broad flow marginal, the frozen field gives an interpolative and locally meaningful target. This is why the main method can query fields on student-rolled states, while dense-query variants still require correlation analysis.

\vspace{0.2cm}

\subsection{Hard-Routed Estimator and Field-Conflict Bias}
\label{app:hard-routing-math}

The hard-routed objective in Sec.~\ref{sec:hard-routing} can be written with $\pi_m=\pi(m)$ as follows:
\begin{equation}
  \mathcal{L}_{\mathrm{route}}
  = \sum_{m=1}^{M}\pi_m\,
  \mathbb{E}_{(x,c)\sim\mathcal{D}_m,\,z_T\sim p_T,\,s\sim q_{\mathrm{sem}}}
  \ell_m(x,c,z_T,s),
  \qquad
  \ell_m(x,c,z_T,s)=\left\|v_\theta(z_t^\theta,t,c)-v_m(z_t^\theta,t,c)\right\|_2^2,\quad t=t(s).
  \label{eq:app-route-obj}
\end{equation}
Here $z_t^\theta=\Phi_\theta^{t(s)}(z_T,c)$ is the stop-gradient student-rolled query state.
Sampling one route $m\sim\pi$ and one sample from $\mathcal{D}_m$ gives an unbiased stochastic estimator of Eq.~\eqref{eq:app-route-obj}. This is the reason the main algorithm does not need to query all fields at every step: the long-run optimization still covers all capabilities, while each individual sample keeps a single semantic target. This corresponds to the first query choice in the main text. The route decides which frozen capability field supervises the current sample, so the target remains a well-defined capability query rather than an average of several fields.

Let $y$ denote the route label of the current sample. A within-sample mixture with nonnegative weights $w_m$, in contrast, replaces the routed target $v_y$ by:
\begin{equation}
  \bar{v}=\sum_{m=1}^{M}w_m v_m,
  \qquad w_m\ge0,\quad \sum_m w_m=1.
\end{equation}
The target bias relative to the correct field is:
\begin{equation}
  \bar{v}-v_y=\sum_{m\ne y}w_m\,(v_m-v_y).
  \label{eq:app-mix-bias}
\end{equation}
When non-route fields encode unrelated or conflicting capabilities, Eq.~\eqref{eq:app-mix-bias} injects irrelevant directions into the sample. This is the algebraic form of target-field ambiguity. Even if the mixed target is numerically valid, its direction need not correspond to the capability associated with the current sample. 

Same-step accumulation avoids this target bias but still averages gradients from $G$ capability buckets in one optimizer update. Let $g_r$ denote the gradient from the $r$-th routed bucket; then:
\begin{equation}
  g_{\mathrm{acc}}=\frac{1}{G}\sum_{r=1}^{G}g_r,
  \qquad
  \|g_{\mathrm{acc}}\|^2=\frac{1}{G^2}\sum_{i=1}^{G}\sum_{j=1}^{G}\langle g_i,g_j\rangle.
  \label{eq:app-grad-conflict}
\end{equation}
If cross-field gradient inner products are small or negative, the same-step update becomes a compromise direction. Here $G$ matches the capability-bucket count used in the diagnostic tables. This is the mathematical form of the capability-field conflict tested in Sec.~\ref{sec:multi-teacher-ext}, and is the same phenomenon targeted by gradient-balancing, multi-objective, gradient surgery, and conflict-averse multi-task optimization methods~\cite{chen2018gradnorm,sener2018multi,yu2020gradient,liu2021conflict,javaloy2021rotograd}.

\vspace{0.2cm}

\subsection{Semantic-Side Query Distribution and Dense Aggregation}
\label{app:query-impl}

Let $s\in[0,1]$ denote a semantic-side coordinate, with larger $s$ closer to the low-noise, clean-image side of the trajectory. The main method samples one query with positive Beta parameters $\alpha_{\mathrm{sem}}$ and $\beta_{\mathrm{sem}}$:
\begin{equation}
  s\sim \mathrm{Beta}(\alpha_{\mathrm{sem}},\beta_{\mathrm{sem}}),
  \qquad \alpha_{\mathrm{sem}}>\beta_{\mathrm{sem}},
  \label{eq:app-beta-query}
\end{equation}
then maps $s$ to the corresponding solver state $z_t^\theta$. Median- and high-noise query ablations are obtained by changing the Beta parameters or using a uniform distribution. The role of Eq.~\eqref{eq:app-beta-query} is not to tune a sampler arbitrarily, but to query a region where capability-specific information is dense. This gives the implementation form of the semantic-side single query used in the main method. The query is placed in the low-noise region where edit, style, and visual-attribute information is more concentrated, while still using the current student rollout state. 

Table~\ref{tab:beta-query-map} spells out the convention used in the ablation table: the Beta random variable is sampled over the normalized rollout index, or semantic coordinate $s$, not directly over the physical flow time $t$.

\begin{table}[t]
  \centering
  \footnotesize
  \setlength{\tabcolsep}{3pt}
  \caption{\textbf{Timestep-Query Distributions.} The rollout index increases from the high-noise beginning of the reverse trajectory to the low-noise semantic side. Therefore, a distribution with a larger mean in $s$ selects later rollout states, which correspond to lower physical noise time.}
  \label{tab:beta-query-map}
  \vspace{-0.2cm}
  \begin{tabular}{@{}p{0.18\linewidth}p{0.19\linewidth}p{0.12\linewidth}p{0.18\linewidth}p{0.23\linewidth}@{}}
    \toprule
    \textbf{Ablation Label} & \textbf{Distribution Over $s$} & \textbf{Mean Of $s$} & \textbf{Selected Rollout Region} & \textbf{Interpretation}\\
    \midrule
    Low-$t$ semantic-side & Beta$(5,2)$ & $\frac{5}{5+2}\approx0.714$ & late indices & low physical $t$, near-clean, semantic-rich\\
    Median-$t$ & Beta$(5,5)$ & $0.5$ & middle indices & intermediate-noise query\\
    High-$t$ noise-side & Beta$(2,5)$ & $\frac{2}{2+5}\approx0.286$ & early indices & high physical $t$, noise-side query\\
    \bottomrule
  \end{tabular}
\end{table}

For dense-query diagnostics with $K>1$, the deterministic ODE rows use the student rollout:
\begin{equation}
  z_{i+1}=z_i-\Delta t\,v_\theta(z_i,t_i,c),
  \label{eq:app-dense-ode-rollout}
\end{equation}
and query $K$ states $\{z_{t_k}^\theta\}_{k=1}^{K}$ from this trajectory. Let:
\begin{equation}
  \ell_k=\left\|v_\theta(z_{t_k}^\theta,t_k,c)-v_m(z_{t_k}^\theta,t_k,c)\right\|_2^2
  \label{eq:app-dense-perstate-loss}
\end{equation}
be the routed velocity loss at the $k$-th queried state. We consider generic aggregations, where $a_k$ denotes weighted-aggregation coefficients, $d_k$ denotes a detached loss value, and $\epsilon_{\mathrm{stab}}>0$ is a numerical stabilizer:
\begin{align}
  \mathcal{L}_{\mathrm{mean}} &= \frac{1}{K}\sum_{k=1}^{K}\ell_k,\\
  \mathcal{L}_{\mathrm{sum}}  &= \sum_{k=1}^{K}\ell_k,\\
  \mathcal{L}_{\mathrm{wtd}}  &= \sum_{k=1}^{K}a_k\ell_k,\\
  a_k &=\frac{d_k}{\sum_{j=1}^{K}d_j+\epsilon_{\mathrm{stab}}},\\
  d_k &=\mathrm{sg}(\ell_k),
  \label{eq:app-dense-weighted}
\end{align}
Here $a_k\ge0$ and $\sum_k a_k\approx1$. Mean aggregation keeps the loss scale fixed but dilutes each state; sum aggregation preserves per-state gradient magnitude but is more sensitive to correlated residuals; weighted aggregation gives higher-disagreement states more credit while avoiding an uncontrolled raw sum. 

In Fig.~\ref{fig:routing-dense-diagnostics} (a), ``isolation'' denotes a $G{=}1$ hard-routing control that increases $K$ without same-step multi-teacher accumulation, while ``sum no control'' denotes Eq.~\eqref{eq:app-dense-perstate-loss} aggregated by $\mathcal{L}_{\mathrm{sum}}$ under ODE rollout, without weighting or SDE decorrelation. These are implementation choices for diagnostics, not the default method.

\vspace{0.2cm}

\subsection{Loss Variants Considered}
\label{app:loss-variants}

Let $v_m$ denote the hard-routed teacher velocity for the current sample, and let $z_t^\theta$ be the stop-gradient student-rolled query state. Throughout this subsection, unadorned velocity terms are evaluated at $(z_t^\theta,t,c)$ unless their arguments are shown explicitly. The default objective is direct velocity regression, that is:
\begin{equation}
  \mathcal{L}_{\mathrm{MSE}}
  = \left\|v_\theta(z_t^\theta,t,c)-v_m(z_t^\theta,t,c)\right\|_2^2.
  \label{eq:app-loss-mse}
\end{equation}
The objective ablations in Fig.~\ref{fig:routing-dense-diagnostics} (c) use the same $K{=}1$, $G{=}1$ single-query ODE setting unless the row explicitly changes the teacher schedule or dense-query setting.

\noindent\textbf{Timestep-Weighted and KL-Weighted MSE.}
We test two weighted velocity-regression forms, using a Min-SNR-style weight $w_{\mathrm{SNR}}(t)$ and local transition variance $\bar{\sigma}_t^2$:
\begin{align}
  \mathcal{L}_{\mathrm{SNR}}
  &= w_{\mathrm{SNR}}(t)
  \left\|v_\theta-v_m\right\|_2^2,\\
  \mathcal{L}_{\mathrm{KL\text{-}\bar{\sigma}^2}}
  &= \frac{\Delta t^2}{2\bar{\sigma}_t^2}
  \left\|v_\theta-v_m\right\|_2^2.
  \label{eq:app-loss-weighted-mse}
\end{align}
The first objective uses timestep weighting; the second follows the Gaussian local-transition KL view in Sec.~\ref{app:kl-mse}. Both keep the same deterministic teacher target as Eq.~\eqref{eq:app-loss-mse}.

\noindent\textbf{DMD-EMA Hybrid and Pure DMD.}
Let $v_{\mathrm{ema}}$ be a passive EMA reference velocity field, and let $\beta\in[0,1]$ be the MSE-anchor weight. The DMD-EMA hybrid uses a score-style surrogate plus an MSE anchor, that is:
\begin{equation}
  \mathcal{L}_{\mathrm{DMD\text{-}EMA}}
  =-(1-\beta)\left\langle \bigl(v_m-v_{\mathrm{ema}}\bigr)_{\mathrm{sg}},v_\theta\right\rangle
  +\beta\left\|v_\theta-v_m\right\|_2^2.
  \label{eq:app-loss-dmd-ema}
\end{equation}
Pure DMD is the special case $\beta=0$:
\begin{equation}
  \mathcal{L}_{\mathrm{pure\text{-}DMD}}
  =-\left\langle \bigl(v_m-v_{\mathrm{ema}}\bigr)_{\mathrm{sg}},v_\theta\right\rangle.
  \label{eq:app-loss-pure-dmd}
\end{equation}
This removes the velocity-regression anchor and is therefore a stability diagnostic rather than a proposed replacement.

\noindent\textbf{SDS+DMD-EMA Combined.}
The combined score-style row applies timestep weighting to the score surrogate and keeps the MSE anchor, that is:
\begin{equation}
  \mathcal{L}_{\mathrm{SDS+DMD}}
  =-(1-\beta)\,w_{\mathrm{SNR}}(t)
  \left\langle \bigl(v_m-v_{\mathrm{ema}}\bigr)_{\mathrm{sg}},v_\theta\right\rangle
  +\beta\left\|v_\theta-v_m\right\|_2^2.
  \label{eq:app-loss-sds-dmd}
\end{equation}
It tests whether timestep weighting and a score-style direction are complementary.

\noindent\textbf{DMD2-Inter and DMD2-Dist.}
The DMD2 rows replace the passive EMA field with a trained fake reference critic $v_{\mathrm{fake}}$ and use mixing weight $\beta_2\in[0,1]$, that is:
\begin{equation}
  \mathcal{L}_{\mathrm{DMD2}}
  =-(1-\beta_2)\left\langle \bigl(v_m-v_{\mathrm{fake}}\bigr)_{\mathrm{sg}},v_\theta\right\rangle
  +\beta_2\left\|v_\theta-v_m\right\|_2^2.
  \label{eq:app-loss-dmd2}
\end{equation}
The DMD2-inter and DMD2-dist rows in Fig.~\ref{fig:routing-dense-diagnostics} (c) are implementation variants of this same score-surrogate family inspired by adversarial and distribution matching distillation~\cite{sauer2024adversarial,sauer2024fast,yin2024one}: DMD2-inter trains the fake critic on rollout-intermediate targets, whereas DMD2-dist trains it on fresh student-distribution denoising targets.

\noindent\textbf{Consistency Variant.}
For two nearby queried states $z_{t_i}$ and $z_{t_j}$, we form a linear clean-image estimate $\hat{x}_{0,a}(z_t)=z_t-t\,v_a(z_t,t,c)$ for $a\in\{\theta,m\}$ and use nonnegative weights $\lambda_{\mathrm{self}}$ and $\lambda_{\mathrm{anchor}}$:
\begin{equation}
  \mathcal{L}_{\mathrm{cons}}
  = \lambda_{\mathrm{self}}
  \left\|\hat{x}_{0,\theta}(z_{t_i})-
      \mathrm{sg}\!\left(\hat{x}_{0,\theta}(z_{t_j})\right)\right\|_2^2
  +\lambda_{\mathrm{anchor}}
  \left\|\hat{x}_{0,\theta}(z_{t_i})-
      \hat{x}_{0,m}(z_{t_i})\right\|_2^2.
  \label{eq:app-loss-consistency}
\end{equation}
This is a simplified consistency regularizer~\cite{song2023consistency,kim2024consistency} with a teacher anchor.

\noindent\textbf{Auxiliary Feature Distillation.}
Let $h_\theta^\ell$ and $h_m^\ell$ be matched student and teacher hidden features at selected DiT layers $\ell\in\mathcal{S}$, and let $\lambda_{\mathrm{feat}}\ge0$ be the feature-distillation weight, following the broader practice of hidden-feature, attention, and representation-level distillation~\cite{romero2014fitnets,zagoruyko2016paying,heo2019comprehensive,ahn2019variational}. The AuxFeat row adds a feature-level term to the velocity MSE, that is:
\begin{equation}
  \mathcal{L}_{\mathrm{AuxFeat}}
  =\left\|v_\theta-v_m\right\|_2^2
  +\lambda_{\mathrm{feat}}\frac{1}{|\mathcal{S}|}
  \sum_{\ell\in\mathcal{S}}\left\|h_\theta^\ell-\mathrm{sg}(h_m^\ell)\right\|_2^2.
  \label{eq:app-loss-auxfeat}
\end{equation}

\noindent\textbf{Soft-Teacher Mixing Objectives.}
The hard-routed KL-$\bar{\sigma}^2$ row and the soft-teacher rows change different axes. KL-$\bar{\sigma}^2$ keeps sample-wise routing and changes only the timestep weighting of a single routed teacher target. Soft-teacher mixing removes routing: for the same student-rolled state, all capability fields are queried and combined inside the same sample objective, similar in spirit to multi-teacher aggregation but without sample-level task selection~\cite{shen2019customizing}.

The soft-teacher MSE row uses nonnegative mixture weights $w_m$:
\begin{equation}
  \mathcal{L}_{\mathrm{soft\text{-}MSE}}
  =\sum_{m=1}^{M}w_m\left\|v_\theta(z_t^\theta,t,c)-v_m(z_t^\theta,t,c)\right\|_2^2,
  \qquad w_m\ge0,\quad \sum_{m=1}^{M}w_m=1.
  \label{eq:app-loss-soft-teacher}
\end{equation}
Equivalently, with $\bar{v}=\sum_m w_m v_m$, that is:
\begin{equation}
  \sum_m w_m\|v_\theta-v_m\|_2^2
  =\|v_\theta-\bar{v}\|_2^2+
    \sum_m w_m\|v_m-\bar{v}\|_2^2.
  \label{eq:app-soft-mse-equivalence}
\end{equation}
Therefore, this has the same gradient as MSE to the mixed velocity target $\bar{v}$. It is also a KL-equivalent quadratic under the equal-covariance Gaussian view of Sec.~\ref{app:kl-mse}; the important difference from KL-$\bar{\sigma}^2$ is the non-routed all-teacher target, not a different Gaussian algebra.

The soft-teacher KL-$\bar{\sigma}^2$ diagnostic applies the same local-transition weighting as the hard-routed KL-$\bar{\sigma}^2$ row to the soft all-teacher objective:
\begin{equation}
  \mathcal{L}_{\mathrm{soft\text{-}KL\text{-}\bar{\sigma}^2}}
  =\frac{\Delta t^2}{2\bar{\sigma}_t^2}
    \sum_{m=1}^{M}w_m\left\|v_\theta(z_t^\theta,t,c)-v_m(z_t^\theta,t,c)\right\|_2^2.
  \label{eq:app-loss-soft-kl-sigma}
\end{equation}
Together, the hard versus soft rows and the MSE versus KL-$\bar{\sigma}^2$ rows form the routing and objective $2\times2$ summarized in panel b of Fig.~\ref{fig:routing-dense-diagnostics}.

These variants correspond to timestep-weighted KL and MSE matching, SNR-style weighting~\cite{hang2023efficient}, score-form surrogate gradients inspired by score distillation and distribution matching~\cite{poole2022dreamfusion,sauer2024adversarial,sauer2024fast,yin2024one}, consistency regularization~\cite{song2023consistency,kim2024consistency}, feature distillation, and soft multi-teacher matching. They are useful ablations, but the unweighted MSE remains our default because it directly matches deterministic capability velocities and empirically gives the strongest stability and performance trade-off.

\noindent\textbf{Implementation Audit.}
Several alternative losses are useful to test, but should be interpreted carefully in our setting. First, the score-distillation-style variants operate on stop-gradient query states; there is no differentiable rendering or sampling path through which an SDS gradient can act on the query state, unlike reward- or preference-optimized diffusion tuning methods that explicitly optimize through policy-gradient, reward-backpropagation, or preference objectives~\cite{black2024training,fan2023dpok,prabhudesai2023aligning,wallace2024diffusion}. Second, the DMD-EMA variant uses a passive EMA reference rather than a separately trained fake-distribution critic. Without the MSE anchor, the score-style gradient need not vanish when the student already matches the routed field, which explains the observed instability of the pure DMD variant. Third, the consistency variant is a simplified diagnostic: it uses a linear $\hat{x}_0$ estimate and a stop-gradient current-student target instead of a full EMA consistency model. 

These audits are the reason we present the non-MSE objectives as ablations rather than as replacements for routed velocity regression.

\vspace{0.2cm}

\subsection{Guidance-Field Absorption}
\label{app:cfg-absorption}

Classifier-free guidance forms an affine velocity field from the unconditional velocity $v_{\emptyset}$ and conditional velocity $v_{\mathrm{cond}}$~\cite{ho2022classifier}:
\begin{equation}
  v_{\alpha}(z_t,t,c)
  =
  v_{\emptyset}(z_t,t)
  +\alpha\bigl(v_{\mathrm{cond}}(z_t,t,c)-v_{\emptyset}(z_t,t)\bigr).
  \label{eq:app-cfg-alpha}
\end{equation}
DanceOPD can treat Eq.~\eqref{eq:app-cfg-alpha} as just another capability field and match it by MSE. In the realizable case, the minimizer of the absorbed-CFG loss
\begin{equation}
  \mathcal{L}_{\mathrm{CFG}}
  =\mathbb{E}\left\|v_\theta(z_t^\theta,t,c)-v_{\alpha}(z_t^\theta,t,c)\right\|_2^2
  \label{eq:app-cfg-mse}
\end{equation}
is $v_\theta=v_{\alpha}$ on the queried state distribution. The guidance direction is then absorbed into the trained student, so the default deployment can use the distilled field directly rather than recomputing the affine extrapolation.

This also clarifies an implementation pitfall. If a model has already learned an $\alpha$-guided conditional branch, but inference applies classifier-free guidance with scale $\beta$ again, then under the approximation $v_{\theta,\emptyset}\approx v_{\emptyset}$ and $v_{\theta,\mathrm{cond}}\approx v_{\emptyset}+\alpha(v_{\mathrm{cond}}-v_{\emptyset})$, the effective field becomes:
\begin{equation}
  v_{\mathrm{eval}}
  =
  v_{\theta,\emptyset}
  +\beta\bigl(v_{\theta,\mathrm{cond}}-v_{\theta,\emptyset}\bigr)
  \approx
  v_{\emptyset}
  +\alpha\beta\bigl(v_{\mathrm{cond}}-v_{\emptyset}\bigr).
  \label{eq:app-cfg-compose}
\end{equation}
Thus, train-time guidance absorption and inference-time CFG scales compose multiplicatively. We include CFG absorption as an optional capability-field construction, not as the default objective for the main experiments.

\vspace{0.2cm}

\subsection{Dense-Query Correlation}
\label{app:dense-correlation}

Suppose a dense trajectory objective queries $K$ states from one student rollout and yields per-query gradient residuals $b_1,\ldots,b_K$. If the residuals were independent, averaging $K$ queries would reduce variance by a factor of $K$. Along one rollout, however, residuals share the same noise seed, condition, student dynamics, and numerical path history. Let $\mathrm{Var}(b_i)=\sigma_b^2$ and $\mathrm{Corr}(b_i,b_j)=\rho$ for $i\ne j$. Then:
\begin{equation}
  \mathrm{Var}\!\left(\frac{1}{K}\sum_{i=1}^K b_i\right)
  =
  \frac{\sigma_b^2}{K}\bigl(1+(K-1)\rho\bigr).
  \label{eq:app-corr-var}
\end{equation}
When $\rho\approx0$, dense queries behave like independent supervision. When $\rho\approx1$, the variance reduction nearly disappears. Worse, if several conflicting capability fields are accumulated in the same optimizer update, correlated queries can repeatedly amplify the same conflicting direction. This explains why dense querying is not automatically superior to a single semantic-side query. This connects the dense-query ablation to the third query choice in the main text. Increasing $K$ changes the number of correlated states from the same rollout, whereas the default method sets $K{=}1$ to keep one teacher query per sample.

\vspace{0.2cm}

\subsection{SDE Decorrelation}
\label{app:sde-decorrelation}

To test whether trajectory correlation is the failure mode, we inject stochasticity into the student rollout. Let $\sigma_i$ be the scheduler noise scale at step $i$, let $\eta$ control the injected stochasticity, and draw $\epsilon_i\sim\mathcal{N}(0,I)$:
\begin{equation}
  z_{i+1}=z_i-\Delta t\,v_\theta(z_i,t_i,c)+\eta\,\sigma_i\sqrt{\Delta t}\,\epsilon_i,
  \qquad \epsilon_i\sim\mathcal{N}(0,I).
  \label{eq:app-sde-rollout}
\end{equation}
This is an Euler Maruyama-style rollout, analogous to the stochastic processes used in score-based generative modeling~\cite{song2020score}. Under a Lipschitz velocity field, independent noise injections perturb nearby rollout states by independent components. Consequently, residual correlations decay with temporal separation; informally, for constants $C_0,\kappa>0$ depending on the local Lipschitz and noise scales:
\begin{equation}
  \left|\mathrm{Corr}(b_i,b_j)\right|
  \le C_0\exp\bigl(-\kappa\eta^2 |i-j|\bigr),
  \label{eq:app-corr-decay}
\end{equation}
We use this only as a diagnostic mechanism: if SDE rollout rescues dense-query accumulation, it supports the hypothesis that correlated trajectory queries, rather than on-policy distillation itself, caused the degradation.

\vspace{0.2cm}

\section{Implementations}
This section covers additional details regarding implementations.

\vspace{0.2cm}

\subsection{Computational Complexity}
\label{app:compute-complexity}

We report the wall-clock cost in Fig.~\ref{fig:overview} as seconds per training step under the same hardware setting. Table~\ref{tab:compute-complexity} decomposes the dominant per-step costs. 

The main algorithmic difference is whether a method first performs an on-policy student rollout, and how many rollout states contribute differentiable training signals. Let $N$ denote the training rollout length, $K$ the number of queried states used in the loss, $B_{\mathrm{phys}}$ the physical micro-batch size, $G_{\mathrm{grp}}$ the FlowGRPO group size, and $\gamma$ the resulting micro-batch factor. We use $C_{\mathrm{roll}}$ for the cost of one no-gradient student rollout step and $C_{\mathrm{grad}}$ for one queried-state supervision unit, including the frozen teacher or reward query and the student forward and backward computation. In the timing setting of Fig.~\ref{fig:overview}, $N{=}16$, $B_{\mathrm{phys}}{=}8$, $G_{\mathrm{grp}}{=}16$, $K_{\mathrm{off}}{=}K_{\mathrm{ours}}{=}1$, and $K_{\mathrm{dense}}{=}N$; therefore $\gamma_{\mathrm{std}}{=}1$ for methods without algorithmic grouping and $\gamma_{\mathrm{flow}}{=}\lceil \frac{G_{\mathrm{grp}}}{B_{\mathrm{phys}}}\rceil{=}2$ for Flow-OPD. The rollout states are stop-gradient states; thus, $K$ counts local gradient-bearing evaluations, not backpropagation through the numerical solver.

\begin{table}[t]
  \centering
  \footnotesize
  \caption{\textbf{Configuration of Per-Step Computational Complexity.} All on-policy variants use the same $N$-step training rollout. The batch factor records extra micro-batches required by algorithmic grouping when the group does not fit in one physical micro-batch.}
  \label{tab:compute-complexity}
  \resizebox{\linewidth}{!}{%
  \begin{tabular}{@{}lccccl@{}}
    \toprule
    \textbf{Method} & \textbf{Query State} & \textbf{Rollout} & \textbf{$K$} & \textbf{Batch Factor} & \textbf{Dominant Per-Step Cost}\\
    \midrule
    Off-Policy & offline noised state & none & $K_{\mathrm{off}}$ & $\gamma_{\mathrm{std}}$ & $K_{\mathrm{off}}C_{\mathrm{grad}}$\\
    DiffusionOPD & student rollout & $N$-step ODE & $K_{\mathrm{dense}}$ & $\gamma_{\mathrm{std}}$ & $N C_{\mathrm{roll}}+K_{\mathrm{dense}}C_{\mathrm{grad}}$\\
    Flow-OPD & student rollout group & $N$-step SDE & $K_{\mathrm{dense}}$ & $\gamma_{\mathrm{flow}}$ & $\gamma_{\mathrm{flow}}(N C_{\mathrm{roll}}+K_{\mathrm{dense}}C_{\mathrm{grad}})$ plus PPO \& log-prob overhead\\
    \name{} & student rollout & $N$-step ODE & $K_{\mathrm{ours}}$ & $\gamma_{\mathrm{std}}$ & $N C_{\mathrm{roll}}+K_{\mathrm{ours}}C_{\mathrm{grad}}$\\
    \bottomrule
  \end{tabular}}
\end{table}

This decomposition explains the training-time pattern in Fig.~\ref{fig:overview}. First, \name{}, DiffusionOPD, and Flow-OPD are all on-policy: with $N{=}16$ in our timing setting, they must roll out a complete $N$-step student trajectory before querying the teacher or reward signal. Off-policy distillation avoids this rollout and is therefore cheap per step, but it queries fixed noised endpoint states rather than the states visited by the current student. 

Second, after the shared rollout, our model uses $K_{\mathrm{ours}}{=}1$: it samples one semantic-side state and accumulates one local velocity-MSE gradient. DiffusionOPD instead supervises the full trajectory with $K_{\mathrm{dense}}{=}N{=}16$, so the expensive teacher-query and backward part is multiplied by the rollout length. 

Third, Flow-OPD uses the same dense $K_{\mathrm{dense}}$ trajectory signal but with FlowGRPO group size $G_{\mathrm{grp}}{=}16$. Since the physical batch size is $B_{\mathrm{phys}}{=}8$, one algorithmic group cannot fit in a single micro-batch and is split into $\gamma_{\mathrm{flow}}{=}2$ micro-batches. This approximately doubles the wall-clock cost relative to the DiffusionOPD reproduction, with additional smaller overhead from SDE sampling, cached log-probabilities, and PPO clipping~\cite{schulman2017proximal,liu2026flow}. Thus, the measured speed order in Fig.~\ref{fig:overview} follows the expected compute hierarchy: off-policy is cheapest but off-distribution, DanceOPD pays the rollout cost but only $K_{\mathrm{ours}}$ gradient state, DiffusionOPD pays dense $K_{\mathrm{dense}}$-state gradients, and Flow-OPD further pays the FlowGRPO micro-batch factor.

\vspace{0.2cm}

\subsection{DiffusionOPD}
\label{app:diffusionopd-reproduction}

Because the DiffusionOPD~\cite{li2026diffusionopd} code was not open-sourced at the time of our experiments, we implemented a paper-faithful reproduction. The key reproduction details are as follows.

We reproduce the second-stage on-policy distillation recipe: task-specific teachers are treated as frozen capability fields, the current student first generates its own stop-gradient trajectory, and teacher supervision is evaluated on the student-visited states. 

We use deterministic ODE rollout as the default DiffusionOPD sampler, following the broader fast-sampling view that diffusion and flow trajectories can be discretized by high-order solver choices~\cite{zhao2023unipc,zhang2022fast,lu2025dpm}. Concretely, the student trajectory is generated by a $16$-step Euler ODE rollout with no SDE noise, matching our 16-step training rollout. Unlike the main \name{} setting, which selects one semantic-side query state, the DiffusionOPD reproduction supervises the full rollout trajectory, i.e., $K{=}16$. 

For a sample routed to teacher $m$, at each visited state $z_{t_j}^{\theta}$, we match the student and teacher transition means. Let $\mu_\theta$ and $\mu_m$ denote the corresponding student and teacher Euler transition means:
\begin{equation}
  \mathcal{L}_{\mathrm{DOPD\text{-}ODE}}
  =\sum_{j=1}^{K}\frac{1}{2}
  \left\|\mu_{\theta}(z_{t_j}^{\theta})-
          \mu_m(z_{t_j}^{\theta})\right\|_2^2,
  \label{eq:app-diffusionopd-replica}
\end{equation}
where $m$ is the routed capability teacher for the current sample. Under Euler flow matching,
$\mu(z_t)=z_t-\Delta t\,v(z_t,t,c)$, so Eq.~\eqref{eq:app-diffusionopd-replica} is equivalent to:
\begin{equation}
  \mathcal{L}_{\mathrm{DOPD\text{-}ODE}}
  =\sum_{j=1}^{K}\frac{\Delta t^2}{2}
  \left\|v_{\theta}(z_{t_j}^{\theta},t_j,c)-
          v_m(z_{t_j}^{\theta},t_j,c)\right\|_2^2.
  \label{eq:app-diffusionopd-velocity}
\end{equation}
This keeps DiffusionOPD's ODE closed-form KL or transition-mean matching form rather than replacing it with our single-query MSE default.

We also reproduce DiffusionOPD's same-step multi-task accumulation at the capability-bucket level. The accumulated losses are averaged before the optimizer step so that the effective learning-rate scale is comparable across blocks.

\vspace{0.2cm}

\subsection{Flow-OPD}
\label{app:flowopd-reproduction}

We adapt the official Flow-OPD~\cite{fang2026flow} code and recipe to our Z-Image setting. The reproduced core is on-policy SDE rollout with cached transition log-probabilities, routed teacher transition-KL rewards, and a PPO-style clipped objective~\cite{schulman2017proximal,liu2026flow}. For each capability bucket, the student samples a global group of $16$ SDE trajectories, uses $K{=}16$ queried states, SDE noise $\eta{=}0.7$, KL scale $-1$, and PPO clip range $10^{-4}$. The Flow-OPD reproduction uses two active capability buckets and merge initialization for both composition blocks.

We disable MAR by setting $\beta{=}0$. Official MAR requires a separate frozen task-agnostic aesthetic teacher LoRA. We do not have a compatible Z-Image MAR teacher; the released Flow-OPD checkpoint is an SD3.5 final student adapter, and the style edit teacher is task-specific rather than a MAR anchor.

\vspace{0.2cm}

\subsection{Off-Policy}
\label{app:offpolicy-reproduction}

The off-policy reproduction keeps the same hard-routed teacher and MSE loss as \name{}, but replaces the student-visited query state with a fixed noising distribution around offline endpoints, following the standard forward noising view of diffusion training~\cite{ho2020denoising}. For an edit sample, the endpoint is the encoded target image; for the prompt-only T2I bucket, no endpoint image is available, so the implementation uses a random latent endpoint. Let $\mathcal{E}$ denote the image encoder:
\begin{equation}
  x^{\mathrm{off}}_{0,m}=
  \begin{cases}
    \mathcal{E}(x_{\mathrm{tgt}}), & m\in\{\mathrm{Edit},\mathrm{Local},\mathrm{Global}\},\\
    \epsilon_0,\ \epsilon_0\sim\mathcal{N}(0,I), & m=\mathrm{T2I}.
  \end{cases}
\end{equation}
We sample a fresh Gaussian noise vector and construct the grid query states using the same scheduler noising rule as SFT, where $\alpha_{t_j}$ and $\sigma_{t_j}$ are scheduler coefficients:
\begin{equation}
  z^{\mathrm{off}}_{t_j}
  =\operatorname{AddNoise}(x^{\mathrm{off}}_{0,m},\epsilon,t_j)
  \equiv \alpha_{t_j}x^{\mathrm{off}}_{0,m}+\sigma_{t_j}\epsilon,
  \qquad \epsilon\sim\mathcal{N}(0,I).
\end{equation}
Let $c_m$ be the route-specific conditioning. The training objective is then plain velocity regression to the routed frozen teacher:
\begin{equation}
  \mathcal{L}_{\mathrm{off}}
  =\mathbb{E}_{m,(x,c),j,\epsilon,\,\epsilon_0}
  \left\|v_{\theta}(z^{\mathrm{off}}_{t_j},t_j,c_m)
  -\mathrm{sg}\!\bigl[v_m(z^{\mathrm{off}}_{t_j},t_j,c_m)\bigr]\right\|_2^2 .
\end{equation}
Thus, this baseline changes only the query-state distribution: $z^{\mathrm{off}}_{t}$ comes from endpoint forward noising rather than the current student rollout $z_t^{\theta}$. For this baseline, we use $K{=}1$, low-$t$ Beta sampling, the same active route set and route probabilities as \name{}, and no KL, SDE, soft-teacher, or extrapolation terms.

\vspace{0.2cm}

\subsection{On-Policy Generative Field Distillation}
\label{sec:settings}

\begin{table}[t]
  \centering
  \setlength{\tabcolsep}{6pt}
  \caption{\textbf{Default Hyper-Parameters.} Unless otherwise stated, all Z-Image \name{} results use these settings.}
  \label{tab:settings}
  \vspace{-0.2cm}
  \begin{tabular}{lc}
    \toprule
    \textbf{Hyper-Parameter} & \textbf{Z-Image \name{}}\\
    \midrule
    Backbone & Z-Image~\cite{cai2025z}\\
    Trainable module & DiT LoRA~\cite{hu2022lora}\\
    LoRA rank & 128\\
    Query state & student rollout state\\
    Rollout steps & 16-step Euler ODE\\
    States per sample $K$ & 1\\
    Route probabilities & uniform over active buckets\\
    Timestep sampling & Beta$(5,2)$ low-$t$ bias\\
    Objective & plain velocity MSE\\
    Extrapolation scale & disabled\\
    Optimizer & AdamW\\
    Learning rate & $2\times10^{-4}$\\
    Gradient accumulation & 4\\
    \bottomrule
  \end{tabular}
\end{table}

\noindent\textbf{Z-Image \name{} Default Configuration.}
Z-Image~\cite{cai2025z} \name{} diagnostics in Fig.~\ref{fig:routing-dense-diagnostics} panels (a) to (c), Fig.~\ref{fig:timestep-init-ablation-curves} panels (a) to (c), and Fig.~\ref{fig:cfg-absorption-map} (b) use the same frozen capability fields unless otherwise stated: attribute and local edit, style and global edit, and the base T2I field. The student is initialized from the attribute-edit SFT LoRA and trained with hard-routed sample-wise velocity matching. Each optimizer step samples one capability route, rolls out the current student for 16 Euler steps, selects one semantic-side low-noise trajectory state, and minimizes plain velocity MSE to the routed frozen field. We disable extrapolation, delta clipping, EMA fake targets, DMD, SDS weighting, and consistency losses in the default setting; those variants are only used in ablations. Table~\ref{tab:settings} lists the default hyper-parameters.

\noindent\textbf{Training Rollout.}
The 16-step rollout above is a stop-gradient query-state generator used during training, not a claim that the deployed sampler must use 16 steps. DanceOPD performs per-state velocity-field matching rather than trajectory-compression distillation, so the training query discretization and the benchmark inference discretization need not be identical~\cite{lu2022dpm,karras2022elucidating}. We use 16 steps for training because it gives on-policy coverage at a manageable cost; evaluations use the benchmark sampler stated in Sec.~\ref{app:evaluation-settings}.

\noindent\textbf{Tab.~\ref{tab:zimage-multi} Details}
For the T2I and Edit Composition block, there are two active buckets: one bucket is T2I, and the other bucket is Edit. The Edit bucket samples from both attribute and style edit data, but both are routed to the same joint Edit teacher; therefore, this setting is not a three-teacher update. For the Local and Global Edit Composition block, there are also two active buckets, with one Local bucket routed to the attribute teacher and one Global bucket routed to the style teacher. Unless otherwise stated, route probabilities are uniform over active buckets; the two main composition blocks therefore use a $1{:}1$ route ratio, and the three-bucket diagnostics involving attribute, style, and T2I fields use a $1{:}1{:}1$ route ratio.

\noindent\textbf{Fig.~\ref{fig:cfg-absorption-map} (c) Details}
The normalized trajectory coordinate is identical across the rollout-step variants: all rows sample $u\sim\mathrm{Beta}(5,2)$, whose mass is biased toward the clean semantic end. The only change is the discretization grid size $N$: after sampling $u$, we quantize it by $\mathrm{idx}=\min(\lfloor uN\rfloor,N-1)$. Thus, a larger $N$ gives a finer low-$t$ grid, but the same clean-side Beta mass is spread across more states, so the probability assigned to the single cleanest state decreases. For example, the cleanest state has probability $\Pr[\mathrm{idx}=N-1]=1-F_{\mathrm{Beta}(5,2)}(1-\frac{1}{N})$. This is $0.167$ for $N{=}8$ because idx $7$ receives $u\in[\frac{7}{8},1]$, but only $0.017$ for $N{=}28$ because idx $27$ receives $u\in[\frac{27}{28},1]$; the remaining low-$t$ mass is distributed over nearby clean-side states such as indices $20$ to $27$. Therefore, increasing the rollout length does not simply sample the cleanest state more often; it refines the clean-side coverage while reducing the mass on the terminal grid point.

\noindent\textbf{Fig.~\ref{fig:cfg-absorption-map} (a)} SD3.5-M exhibits weak baseline realism and thus benefits from more pronounced improvements. The realism teacher is obtained by full-parameter training with SD3.5-M~\cite{esser2024scaling} for $100{,}000$ steps, using a learning rate of $1\times10^{-5}$ and a batch size of $16$. Fig.~\ref{fig:cfg-absorption-map} (a) then reports the subsequent OPD and distillation steps up to $3$k, not the teacher-training step count.
Our reward model is proprietary, which assigns a photorealism score to each input image, and we only use it for evaluation. For reward evaluation, we used $200$ held-out samples excluded from the training data. Inference was performed at $1024\times1024$ resolution with $28$ sampling steps and CFG scale $3.5$.

\noindent\textbf{Other Experiments Details}
For results in Tab.~\ref{tab:ablation-summary} and Tab.~\ref{tab:rollout-step-sensitivity}, we report checkpoints from 500 to 2000 steps. In Tab.~\ref{tab:zimage-multi}, T2I refers to the Z-Image model. Local Edit is obtained by training OmniEdit~\cite{wei2025omniedit} for one epoch on attribute subsets, Global Edit on style subsets, and Edit on the entire dataset.

\vspace{0.2cm}

\subsection{Evaluation Settings}
\label{app:evaluation-settings}

\begin{table}[t]
  \centering
  \scriptsize
  \setlength{\tabcolsep}{3.5pt}
  \caption{\textbf{CFG Absorption Diagnostics on GEditBench-EN.} Training-time CFG scale $\alpha$ and inference-time CFG scale $\beta$ compose approximately as $\alpha\beta$, but overly large training-time guidance can still damage the learned field.}
  \label{tab:cfg-absorption}
  \resizebox{\linewidth}{!}{%
  \begin{tabular}{lcc|cccccc|c}
    \toprule
    \textbf{Train $\alpha$} & \textbf{Eval $\beta$} & \textbf{Eff. $\alpha\beta$} & \textbf{Subj-Add} & \textbf{Subj-Rep} & \textbf{Bg-Chg} & \textbf{Style-Chg} & \textbf{Color-Alt} & \textbf{Subj-Rem} & \textbf{Avg}\\
    \midrule
    $3.5$ & $1.0$ & $3.5$ &  $6.485$ & $5.897$ & $4.824$ & $5.907$ & $5.628$ & $3.790$ & $5.422$\\
    $1.0$ & $7.0$ & $7.0$ &  $6.266$ & $6.181$ & $5.924$ & $5.060$ & $5.716$ & $5.357$ & $5.751$\\
    $3.5$ & $2.0$ & $7.0$ &  $6.553$ & $6.507$ & $5.911$ & $5.814$ & $5.998$ & $4.215$ & $5.833$\\
    $2.0$ & $7.0$ & $14.0$ & $4.684$ & $5.402$ & $4.114$ & $3.750$ & $4.313$ & $5.115$ & $4.563$\\
    $7.0$ & $7.0$ & $49.0$ &  $3.645$ & $4.367$ & $4.090$ & $3.321$ & $3.902$ & $4.765$ & $4.015$\\

    \bottomrule
  \end{tabular}%
  }
\end{table}

\noindent\textbf{GEditBench-EN.}
For image-editing and edit-style capability composition, we follow the GEditBench-EN~\cite{liu2025step1x} protocol and report the six task categories used in the main table: subject addition, subject replacement, background change, style change, color alteration, and subject removal. The reported average is the arithmetic mean over these six categories, aligning with OmniEdit~\cite{wei2025omniedit}'s six categories. 

Broader image-generation evaluation has also studied reference-free caption alignment, question-answering faithfulness, human-preference rewards~\cite{xu2023imagereward,kirstain2023pick,wu2023human}, compositional prompts, perceptual similarity, and multimodal or physical-world reasoning~\cite{hessel2021clipscore,hu2023tifa,huang2025t2i,fu2023dreamsim,chow2025physbench}. Unless otherwise specified, ablation scores are the GEditBench-EN average, because it is the most direct measurement of whether the distilled student preserves the routed editing capability. For the default non-CFG-sweep evaluations, we use 28 flow-matching sampling steps, and CFG scale $7.0$.

\noindent\textbf{GenEval.}
For testing T2I, we report the standard GenEval categories~\cite{ghosh2023geneval}: single object, two objects, counting, colors, position, and color attribution, together with the overall score. Rows whose target capability is not a T2I generation field may still include GenEval when preservation of the base model is part of the claim. For the default non-CFG-sweep evaluations, we use 28 flow-matching sampling steps, and CFG scale $3.5$.

\begin{table}[H]
  \centering
  \scriptsize
  \setlength{\tabcolsep}{2.35pt}
  \caption{\textbf{Objective, Teacher-Schedule, and Dense-Query Diagnostics.} All rows use the same capability fields and report GEditBench-EN sub-scores. The teacher-schedule column isolates how teacher signals enter an optimizer update: step alternation ($G{=}1$) routes one field per step, same-step accumulation ($G{=}3$) averages several routed losses before one update, and soft-teacher mixing is a non-routed objective diagnostic. The objective and query tag column records the loss family and, when applicable, dense-query aggregation and rollout control.}
  \label{tab:multi-teacher-aggregation}
  \resizebox{\textwidth}{!}{%
  \begin{tabular}{lll|ccccccc}
    \toprule
    \textbf{Teacher Schedule} & \textbf{Objective and Query Tag} & \textbf{Config} & \textbf{Subj-Add} & \textbf{Subj-Rep} & \textbf{Bg-Chg} & \textbf{Style-Chg} & \textbf{Color-Alt} & \textbf{Subj-Rem} & \textbf{Avg}\\
    \midrule
    Step alternation & MSE with ODE & $K{=}1$, $G{=}1$ & $6.266$ & $6.181$ & $5.924$ & $5.060$ & $5.716$ & $5.357$ & $5.751$\\
    Step alternation & Timestep-weighted MSE with ODE & $K{=}1$, $G{=}1$ & $5.953$ & $6.151$ & $5.400$ & $3.986$ & $5.334$ & $6.729$ & $5.592$\\
    Step alternation & DMD-EMA hybrid with ODE & $K{=}1$, $G{=}1$ & $5.731$ & $6.219$ & $5.093$ & $4.752$ & $5.558$ & $6.228$ & $5.597$\\
    Step alternation & SDS+DMD-EMA with ODE & $K{=}1$, $G{=}1$ & $6.397$ & $6.032$ & $5.037$ & $4.040$ & $5.289$ & $5.943$ & $5.456$\\
    Step alternation & Consistency with ODE & $K{=}1$, $G{=}1$ & $5.929$ & $6.039$ & $5.117$ & $5.306$ & $4.907$ & $5.843$ & $5.523$\\
    Step alternation & KL-$\bar{\sigma}^2$ with ODE & $K{=}1$, $G{=}1$ & $5.906$ & $6.061$ & $5.436$ & $4.910$ & $5.427$ & $5.268$ & $5.501$\\
    Step alternation & DMD2-inter with ODE & $K{=}1$, $G{=}1$ & $5.526$ & $5.597$ & $4.975$ & $4.883$ & $5.231$ & $3.637$ & $4.975$\\
    Step alternation & DMD2-dist with ODE & $K{=}1$, $G{=}1$ & $5.596$ & $5.856$ & $4.345$ & $3.016$ & $5.060$ & $5.124$ & $4.833$\\
    Step alternation & AuxFeat with ODE & $K{=}1$, $G{=}1$ & $4.864$ & $5.568$ & $4.850$ & $3.885$ & $4.019$ & $5.303$ & $4.748$\\
    \midrule
    Soft-teacher mixing & Soft-teacher MSE with ODE & $K{=}1$, all teachers & $5.410$ & $5.808$ & $4.902$ & $4.426$ & $5.191$ & $4.225$ & $4.994$\\
    Soft-teacher mixing & Soft-teacher KL-$\bar{\sigma}^2$ with ODE & $K{=}1$, all teachers & $5.302$ & $5.626$ & $4.991$ & $4.290$ & $5.042$ & $4.604$ & $4.976$\\
    \midrule
    Same-step accumulation & MSE with ODE & $K{=}1$, $G{=}3$ & $5.170$ & $5.754$ & $5.754$ & $5.427$ & $6.171$ & $4.632$ & $5.485$\\
    Same-step accumulation & MSE with SDE & $K{=}1$, $G{=}3$, $\eta{=}0.3$ & $4.862$ & $5.068$ & $4.358$ & $4.719$ & $4.816$ & $4.048$ & $4.645$\\
    \midrule
    Step alternation & MSE with ODE, sum & $K{=}2$, $G{=}1$ & $6.339$ & $6.075$ & $5.573$ & $3.159$ & $5.792$ & $5.955$ & $5.482$\\
    Step alternation & MSE with SDE, sum & $K{=}2$, $G{=}1$, $\eta{=}0.3$ & $5.366$ & $5.856$ & $4.461$ & $3.882$ & $5.205$ & $5.381$ & $5.025$\\
    Step alternation & MSE with ODE, weighted & $K{=}2$, $G{=}1$ & $4.850$ & $5.873$ & $5.132$ & $4.094$ & $4.522$ & $5.118$ & $4.931$\\
    Step alternation & MSE with ODE, weighted & $K{=}4$, $G{=}1$ & $5.962$ & $5.886$ & $5.500$ & $5.065$ & $5.654$ & $3.912$ & $5.330$\\
    Step alternation & MSE with ODE, isolation & $K{=}4$, $G{=}1$ & $5.476$ & $5.654$ & $4.995$ & $4.540$ & $5.138$ & $4.066$ & $4.978$\\
    Step alternation & MSE with ODE, isolation & $K{=}8$, $G{=}1$ & $6.207$ & $5.473$ & $4.687$ & $3.716$ & $4.953$ & $5.837$ & $5.145$\\
    Step alternation & MSE with ODE, weighted & $K{=}8$, $G{=}1$ & $5.224$ & $5.877$ & $5.572$ & $4.085$ & $4.821$ & $5.728$ & $5.218$\\
    Step alternation & MSE with ODE, sum no control & $K{=}8$, $G{=}1$ & $5.913$ & $5.744$ & $4.620$ & $3.723$ & $4.817$ & $5.409$ & $5.038$\\
    Step alternation & MSE with ODE, weighted & $K{=}16$, $G{=}1$ & $5.638$ & $6.062$ & $4.776$ & $3.878$ & $5.181$ & $5.227$ & $5.127$\\
    \midrule
    Same-step accumulation & MSE with ODE & $K{=}2$, $G{=}3$ & $4.458$ & $5.193$ & $5.265$ & $4.144$ & $4.672$ & $2.894$ & $4.437$\\
    Same-step accumulation & MSE with SDE & $K{=}2$, $G{=}3$, $\eta{=}0.3$ & $5.751$ & $5.994$ & $5.832$ & $4.114$ & $5.153$ & $4.689$ & $5.255$\\
    Same-step accumulation & MSE with SDE & $K{=}16$, $G{=}3$, $\eta{=}0.3$ & $5.093$ & $5.422$ & $4.583$ & $3.029$ & $4.685$ & $5.939$ & $4.792$\\
    \bottomrule
  \end{tabular}%
  }
\end{table}

\begin{table}[H]
  \centering
  \scriptsize
  \setlength{\tabcolsep}{2.7pt}
  \caption{\textbf{Timestep and Initialization Ablations.}
  Scores are GEditBench-EN sub-scores unless otherwise noted. Objective, teacher-schedule, dense-query, and rollout-control diagnostics are consolidated in Table~\ref{tab:multi-teacher-aggregation}. These ablations support semantic-side low-$t$ querying and the default initialization choice.}
  \label{tab:ablation-summary}
  \resizebox{\textwidth}{!}{%
  \begin{tabular}{lll|ccccccc}
    \toprule
    \textbf{Axis} & \textbf{Variant} & \textbf{Step} & \textbf{Subj-Add} & \textbf{Subj-Rep} & \textbf{Bg-Chg} & \textbf{Style-Chg} & \textbf{Color-Alt} & \textbf{Subj-Rem} & \textbf{Avg}\\
    \midrule
    Timestep query & Low-$t$ & 500 & $5.776$ & $5.759$ & $5.371$ & $4.148$ & $5.033$ & $4.728$ & $5.136$\\
    Timestep query & Low-$t$ & 1000 & $5.543$ & $5.725$ & $5.327$ & $4.515$ & $5.445$ & $5.586$ & $5.357$\\
    Timestep query & Low-$t$ & 1500 & $5.233$ & $5.787$ & $5.198$ & $4.394$ & $5.208$ & $5.161$ & $5.163$\\
    Timestep query & Low-$t$ & 2000 & $6.266$ & $6.181$ & $5.924$ & $5.060$ & $5.716$ & $5.357$ & $5.751$\\
    Timestep query & Median-$t$ & 500 & $5.626$ & $5.902$ & $4.862$ & $4.387$ & $4.991$ & $5.781$ & $5.258$\\
    Timestep query & Median-$t$ & 1000 & $5.548$ & $5.779$ & $4.951$ & $4.538$ & $5.383$ & $4.393$ & $5.099$\\
    Timestep query & Median-$t$ & 1500 & $5.405$ & $5.567$ & $4.576$ & $4.914$ & $5.485$ & $4.233$ & $5.030$\\
    Timestep query & Median-$t$ & 2000 & $4.611$ & $5.656$ & $4.353$ & $4.496$ & $5.012$ & $3.764$ & $4.649$\\
    Timestep query & High-$t$ & 500 & $5.698$ & $5.959$ & $5.113$ & $5.359$ & $5.292$ & $5.099$ & $5.420$\\
    Timestep query & High-$t$ & 1000 & $4.524$ & $4.580$ & $4.801$ & $2.642$ & $3.978$ & $3.727$ & $4.042$\\
    Timestep query & High-$t$ & 1500 & $3.473$ & $4.030$ & $3.947$ & $3.129$ & $4.287$ & $3.480$ & $3.724$\\
    Timestep query & High-$t$ & 2000 & $4.290$ & $5.485$ & $5.186$ & $3.221$ & $5.166$ & $5.528$ & $4.813$\\
    \midrule
    Initialization & Merged & 500 & $3.441$ & $4.395$ & $4.752$ & $5.445$ & $5.240$ & $2.002$ & $4.213$\\
    Initialization & Merged & 1000 & $3.113$ & $4.269$ & $3.119$ & $4.008$ & $3.821$ & $1.040$ & $3.228$\\
    Initialization & Merged & 1500 & $4.485$ & $5.279$ & $5.091$ & $5.023$ & $4.999$ & $1.991$ & $4.478$\\
    Initialization & Merged & 2000 & $4.212$ & $4.680$ & $4.603$ & $4.845$ & $4.426$ & $2.392$ & $4.193$\\
    Initialization & T2I & 500 & $1.443$ & $1.933$ & $2.384$ & $2.088$ & $1.316$ & $0.157$ & $1.554$\\
    Initialization & T2I & 1000 & $1.912$ & $2.980$ & $2.869$ & $2.301$ & $2.410$ & $0.166$ & $2.106$\\
    Initialization & T2I & 1500 & $2.006$ & $3.205$ & $2.856$ & $2.290$ & $2.272$ & $0.336$ & $2.161$\\
    Initialization & T2I & 2000 & $1.571$ & $2.338$ & $3.289$ & $2.024$ & $1.837$ & $0.276$ & $1.889$\\
    Initialization & Local Edit & 500 & $5.776$ & $5.759$ & $5.371$ & $4.148$ & $5.033$ & $4.728$ & $5.136$\\
    Initialization & Local Edit & 1000 & $5.543$ & $5.725$ & $5.327$ & $4.515$ & $5.445$ & $5.586$ & $5.357$\\
    Initialization & Local Edit & 1500 & $5.233$ & $5.787$ & $5.198$ & $4.394$ & $5.208$ & $5.161$ & $5.163$\\
    Initialization & Local Edit & 2000 & $6.266$ & $6.181$ & $5.924$ & $5.060$ & $5.716$ & $5.357$ & $5.751$\\
    Initialization & Global Edit & 500 & $3.459$ & $4.670$ & $4.903$ & $5.406$ & $4.348$ & $1.675$ & $4.077$\\
    Initialization & Global Edit & 1000 & $2.922$ & $4.178$ & $4.644$ & $4.563$ & $4.247$ & $0.702$ & $3.543$\\
    Initialization & Global Edit & 1500 & $3.563$ & $4.349$ & $5.132$ & $5.182$ & $3.958$ & $1.246$ & $3.905$\\
    Initialization & Global Edit & 2000 & $2.609$ & $3.887$ & $3.661$ & $3.112$ & $2.643$ & $0.300$ & $2.702$\\
    \bottomrule
  \end{tabular}%
  }

\end{table}

\begin{table}[t]
  \centering
  \vspace{0.2cm}
  \scriptsize
  \setlength{\tabcolsep}{2.2pt}
  \caption{\textbf{Training Rollout-Step Sensitivity.} All rows use hard-routed MSE with ODE with $K{=}1$, $G{=}1$, low-$t$ Beta$(5,2)$ query sampling, and the same evaluation protocol. Only the number of stop-gradient student-rollout steps used to generate training query states is varied. We report both GEditBench-EN and GenEval because rollout discretization may affect edit quality and base T2I preservation differently.}
  \label{tab:rollout-step-sensitivity}
  \resizebox{\textwidth}{!}{%
  \begin{tabular}{ll|ccccccc|c}
    \toprule
    & & \multicolumn{7}{c|}{\textbf{GEditBench-EN}} & \textbf{GenEval}\\
    \textbf{Rollout} & \textbf{Step} & \textbf{Subj-Add} & \textbf{Subj-Rep} & \textbf{Bg-Chg} & \textbf{Style-Chg} & \textbf{Color-Alt} & \textbf{Subj-Rem} & \textbf{Avg} & \textbf{Overall}\\
    \midrule
    $8$  & $500$  & $5.067$ & $5.237$ & $4.372$ & $4.184$ & $4.564$ & $3.611$ & $4.506$ & $0.833$\\
    $8$  & $1000$ & $5.237$ & $5.819$ & $4.735$ & $4.500$ & $4.962$ & $4.271$ & $4.921$ & $0.855$\\
    $8$  & $1500$ & $5.500$ & $5.788$ & $5.559$ & $4.846$ & $5.079$ & $4.642$ & $5.236$ & $0.866$\\
    $8$  & $2000$ & $6.514$ & $6.137$ & $5.532$ & $5.163$ & $6.346$ & $4.744$ & $5.739$ & $0.852$\\
    $16$ & $500$  & $5.776$ & $5.759$ & $5.371$ & $4.148$ & $5.033$ & $4.728$ & $5.136$ & $0.821$\\
    $16$ & $1000$ & $5.543$ & $5.725$ & $5.327$ & $4.515$ & $5.445$ & $5.586$ & $5.357$ & $0.862$\\
    $16$ & $1500$ & $5.233$ & $5.787$ & $5.198$ & $4.394$ & $5.208$ & $5.161$ & $5.163$ & $0.854$\\
    $16$ & $2000$ & $6.266$ & $6.181$ & $5.924$ & $5.060$ & $5.716$ & $5.357$ & $5.751$ & $0.858$\\
    $20$ & $500$  & $6.045$ & $5.678$ & $5.089$ & $4.180$ & $4.838$ & $6.144$ & $5.329$ & $0.832$\\
    $20$ & $1000$ & $5.995$ & $5.660$ & $6.109$ & $4.733$ & $5.833$ & $5.570$ & $5.650$ & $0.855$\\
    $20$ & $1500$ & $5.339$ & $5.603$ & $4.958$ & $5.034$ & $5.237$ & $3.916$ & $5.014$ & $0.846$\\
    $20$ & $2000$ & $5.889$ & $5.899$ & $5.440$ & $5.042$ & $5.696$ & $5.531$ & $5.583$ & $0.842$\\
    $28$ & $500$  & $4.208$ & $5.271$ & $4.338$ & $3.686$ & $4.323$ & $2.779$ & $4.101$ & $0.849$\\
    $28$ & $1000$ & $3.983$ & $5.175$ & $3.996$ & $3.120$ & $4.228$ & $2.527$ & $3.838$ & $0.866$\\
    $28$ & $1500$ & $4.523$ & $5.409$ & $4.699$ & $4.220$ & $4.697$ & $3.157$ & $4.451$ & $0.851$\\
    $28$ & $2000$ & $6.544$ & $6.147$ & $5.618$ & $5.393$ & $6.475$ & $4.008$ & $5.697$ & $0.834$\\
    \bottomrule
  \end{tabular}%
  }
\end{table}

\section{Additional Qualitative Results}
\label{app:additional-qualitative}

\begin{figure}[t]
    \centering
    \vspace{-0.5cm}
    \includegraphics[width=\textwidth]{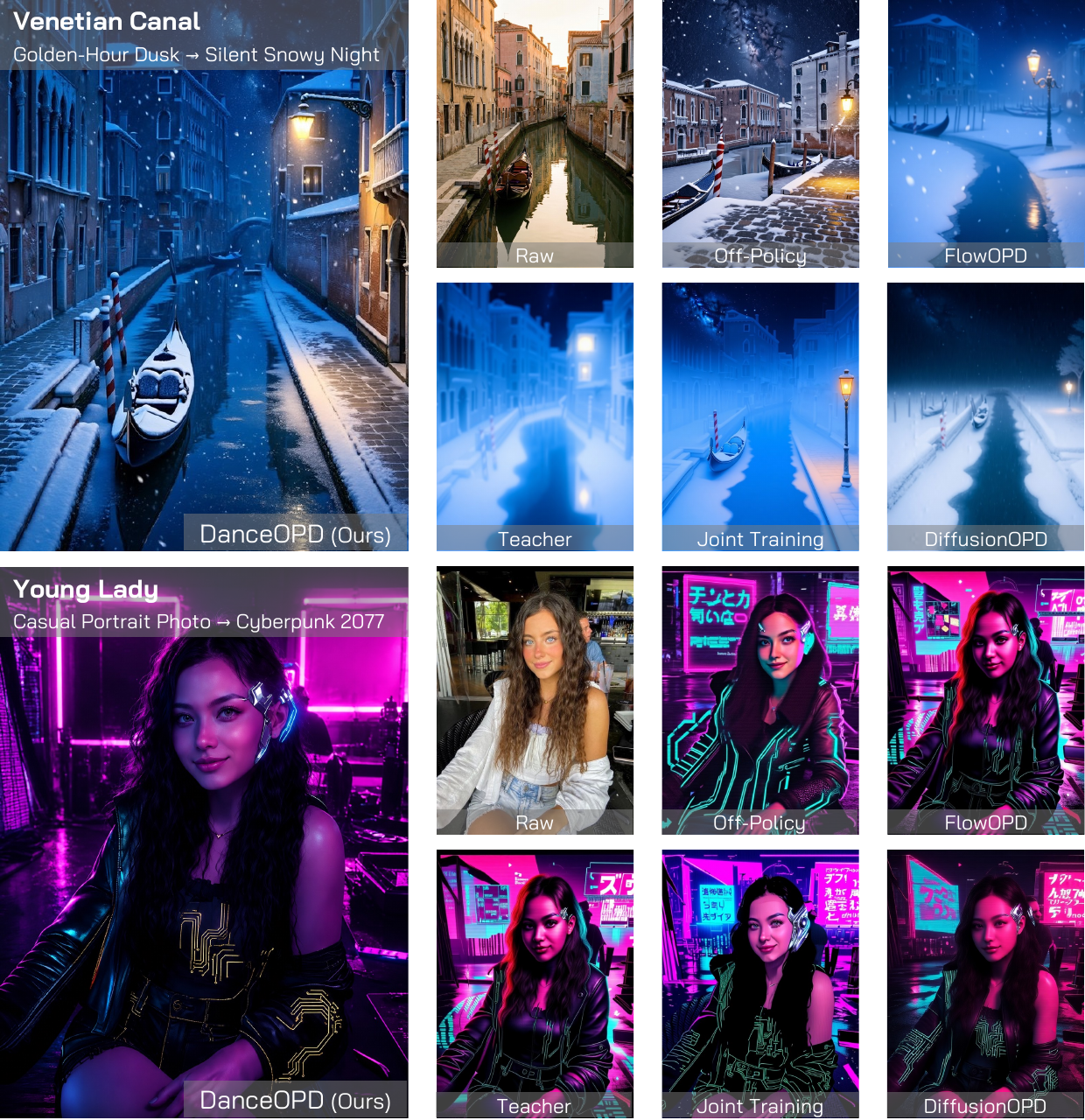}
    \vspace{-0.5cm}
    \caption{\textbf{T2I and Edit Composition: Global Edits.} DanceOPD better follows large style and scene transformations while preserving image structure compared with off-policy distillation, joint training, DiffusionOPD, and Flow-OPD.}
    \label{fig:editing_1}
\end{figure}

\noindent\textbf{Global Scene and Style Edits.}
Fig.~\ref{fig:editing_1} compares large transformation edits where both the global appearance and the original scene identity must be controlled. 

In the Venetian canal example, the requested change is from golden-hour dusk to a silent snowy night. \name{} converts the warm canal scene into a coherent blue night scene with visible snow, snow-covered walkways, and lamp illumination, while preserving the canal geometry and the gondola-like foreground structure. Several baselines either under-transform the scene, introduce excessive blur, or lose part of the original spatial structure. 

In the portrait example, the target cyberpunk style requires strong neon lighting, a darker futuristic environment, and visible local accessories. Our model applies the magenta and cyan cyberpunk appearance while retaining the main face, pose, hair layout, and portrait composition more consistently than the alternatives.

\begin{figure}[t]
    \centering
    \vspace{-0.5cm}
    \includegraphics[width=\textwidth]{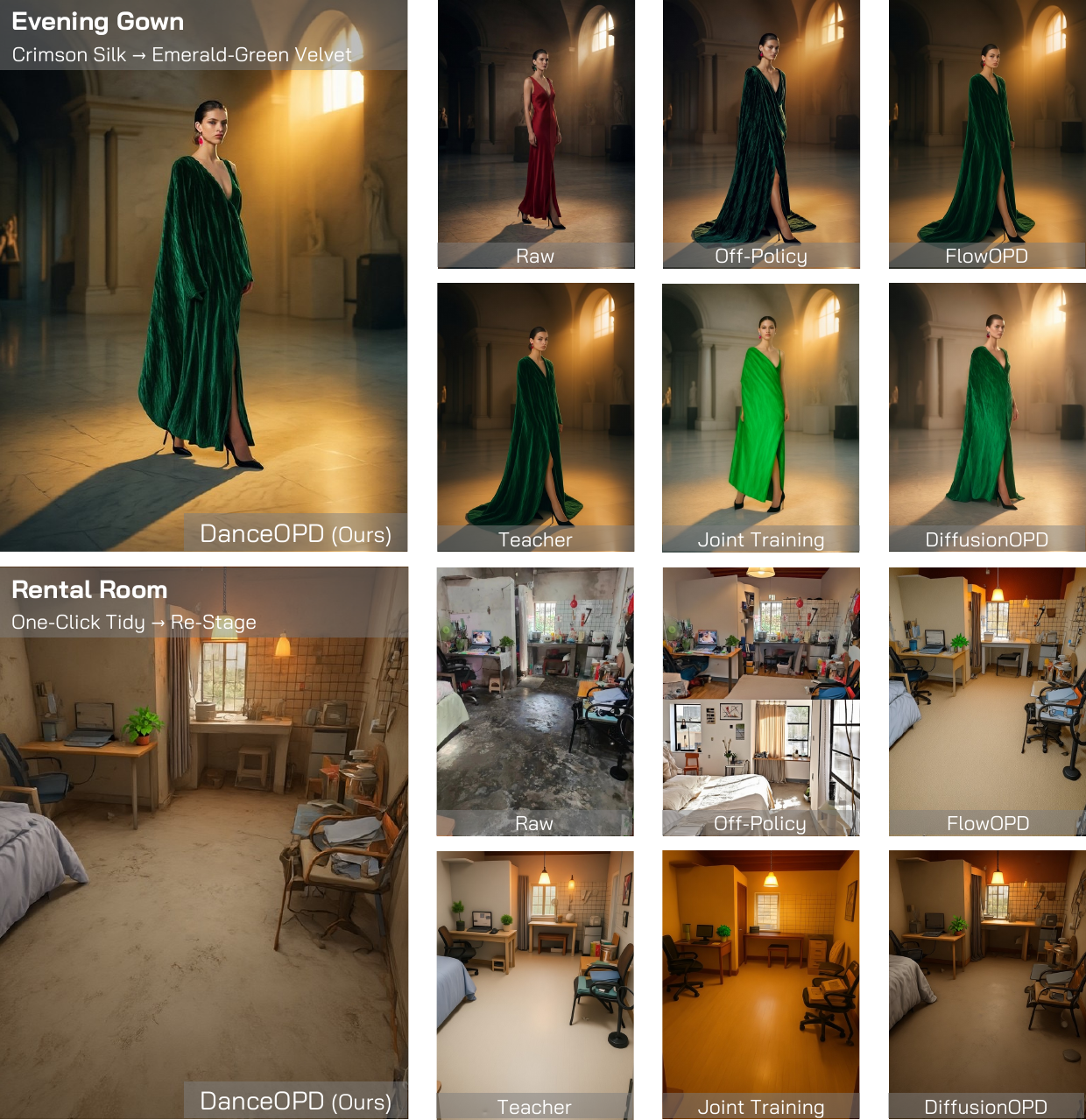}
    \vspace{-0.5cm}
    \caption{\textbf{T2I and Edit Composition: Local and Global Edits.} DanceOPD preserves content under local edits and produces stronger global transformations than the competing baselines.}
    \label{fig:editing_2}
\end{figure}

\noindent\textbf{Local Attribute Changes and Re-Staging.}
Fig.~\ref{fig:editing_2} shows two cases that require different preservation and transformation balances. 

For the evening-gown edit, the main object and the indoor lighting should remain stable, while the dress changes from crimson silk to emerald-green velvet. \name{} changes both the color and material impression of the gown, preserving the human pose and the hall background; in contrast, some baselines either keep a weaker material change, flatten the green dress texture, or alter the image style more globally. 

For the rental-room edit, the target is broader: the room should become tidier and re-staged while still looking like the same indoor space. Our model removes much of the clutter and produces a cleaner room layout without turning the example into an unrelated scene, whereas other methods show stronger layout drift, collage-like structure changes, or incomplete tidying.

\begin{figure}[t]
    \centering
    \vspace{-1.5cm}
    \includegraphics[width=\textwidth]{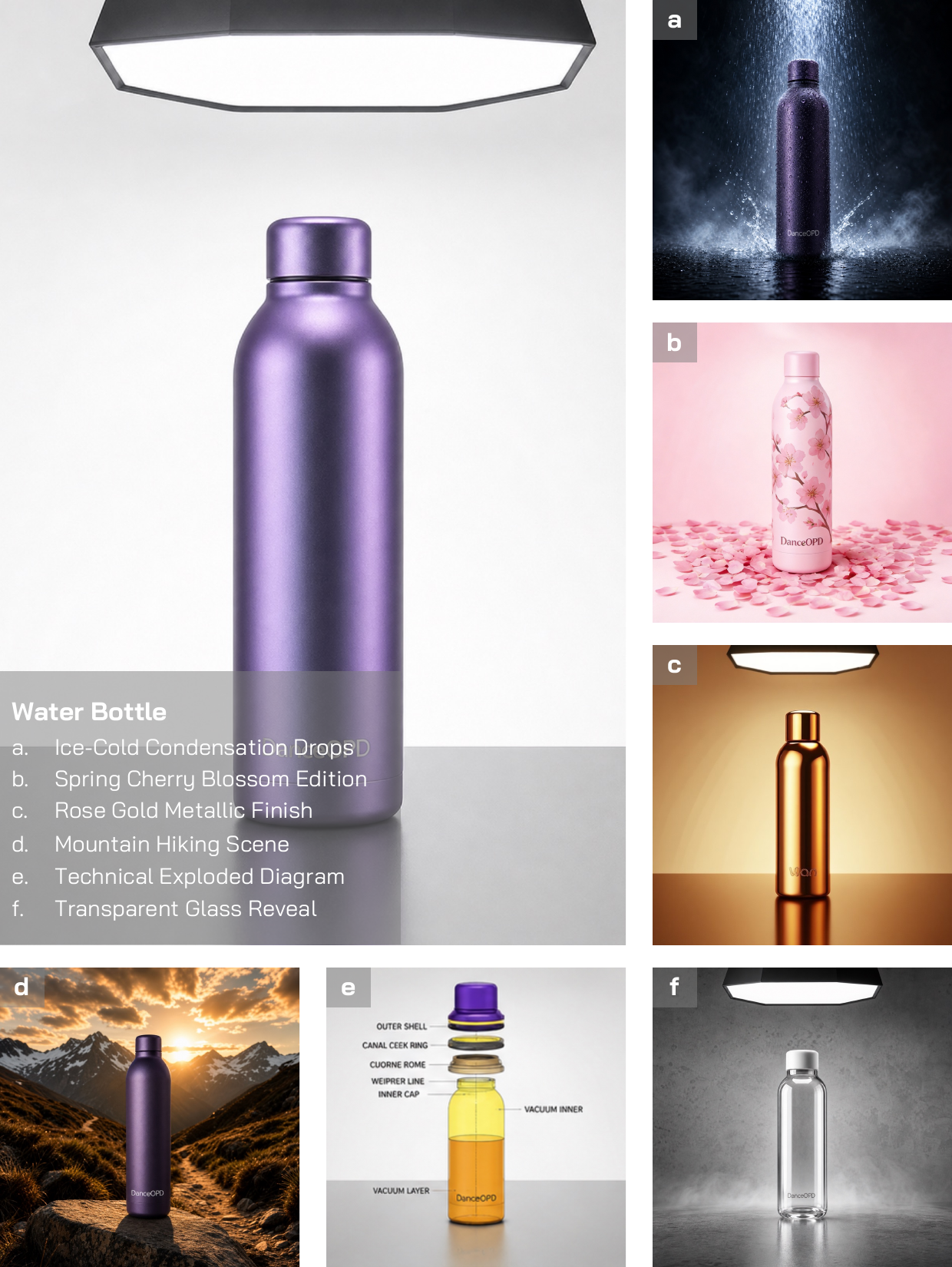}
    \vspace{-0.5cm}
    \caption{\textbf{T2I and Edit Composition: Various Edits on the Same Object.} \name{} supports diverse object-level transformations on the same source object, including material, style, scene, structure, and transparency edits.}
    \label{fig:shared-object-edits}
\end{figure}

\noindent\textbf{Multiple Edits on a Shared Source Object.}
Fig.~\ref{fig:shared-object-edits} edits the same water-bottle source with six different instructions. The results cover local surface changes, such as condensation drops and rose-gold metallic finish; decorative restyling, such as the cherry-blossom edition; context changes, such as placing the bottle in a mountain hiking scene; structural reinterpretation, such as the technical exploded diagram; and material transformation, such as a transparent glass reveal. 

Across these edits, \name{} generally preserves the bottle identity, cap shape, upright product composition, and central placement while changing the requested visual factor, a regime closely related to object binding, structured guidance, and composable-condition control in text-to-image generation~\cite{chefer2023attend,feng2022training,huang2023composer}. This illustrates that the distilled student is not merely memorizing one edit mode: it can reuse the shared object representation across diverse routed capability queries.

\vspace{0.2cm}

\section{Additional Quantitative Results}
\label{app:detailed-results}

Tables~\ref{tab:cfg-absorption} to~\ref{tab:rollout-step-sensitivity} provide the detailed numeric results behind the main CFG, routing, objective, timestep, initialization, and rollout-step analyses.

\noindent\textbf{Diagnostic Scope.}
These tables should be read as controlled diagnostics rather than as additional main-result blocks. Table~\ref{tab:zimage-multi} compares complete composition settings, whereas Tables~\ref{tab:cfg-absorption} to~\ref{tab:rollout-step-sensitivity} isolate individual implementation and estimator choices under matched capability fields. 

Consequently, the same default diagnostic anchor can appear in multiple places: hard-routed MSE with ODE, $K{=}1$, $G{=}1$, a 16-step training rollout, and low-$t$ query sampling. Repeated rows or repeated averages should therefore be interpreted as shared controls for ablations, not as independent duplicates of the Table~\ref{tab:zimage-multi} main settings.

\noindent\textbf{CFG Absorption Diagnostics.}
Table~\ref{tab:cfg-absorption} separates training-time absorbed guidance from inference-time external CFG. The column $\alpha$ denotes the guidance scale used to define the target field during absorption, while $\beta$ denotes the guidance scale applied again at evaluation. 

Under the affine CFG approximation in Sec.~\ref{app:cfg-absorption}, the effective guidance strength is roughly $\alpha\beta$. Thus, $\alpha{=}1,\beta{=}7$ is the eval-only CFG control, $\alpha{=}3.5,\beta{=}1$ measures train-only absorption, and $\alpha{=}3.5,\beta{=}2$ tests a moderate composition of absorbed and external guidance. The large-drop rows illustrate the expected over-guidance failure mode when the effective scale becomes too high.

\noindent\textbf{Rollout-Step Diagnostics.}
Table~\ref{tab:rollout-step-sensitivity} varies only the number of stop-gradient student-rollout steps used to generate training query states. 

The benchmark evaluation sampler is held fixed, so the rows should not be read as evaluating the final model with different sampling-step budgets. Because all rows use the same Beta$(5,2)$ distribution over the normalized semantic coordinate, increasing the rollout length refines the clean-side grid but also spreads the same probability mass over more candidate states. 

This is why longer training rollouts do not automatically improve GEditBench or GenEval: the rollout is a query-state generator for field matching, not a trajectory-compression target.

\clearpage\clearpage
\bibliographystyle{plainnat}
\bibliography{main}

\end{document}